%% file: neurips_2026_final.tex
\documentclass{article}

\usepackage[final, eandd]{neurips_2026}

\usepackage[utf8]{inputenc}
\usepackage[T1]{fontenc}
\usepackage{hyperref}
\usepackage{url}
\usepackage{booktabs}
\usepackage{amsfonts}
\usepackage{nicefrac}
\usepackage{microtype}
\usepackage{xcolor}
\usepackage{graphicx}
\usepackage{subfigure}
\usepackage{enumitem}
\usepackage{wrapfig}
\usepackage{algorithm}
\usepackage{algpseudocode}
\usepackage{amsmath}
\usepackage{amssymb}
\usepackage{colortbl}
\usepackage{float}
\usepackage{mathtools}
\usepackage{multirow}
\usepackage{array}
\usepackage{siunitx}
\usepackage{amsthm}
\usepackage{xcolor}
\usepackage{booktabs}
\usepackage{colortbl}
\usepackage{array}

\definecolor{cat1color}{HTML}{2196F3}
\definecolor{cat2color}{HTML}{FF9800}
\definecolor{cat3color}{HTML}{4CAF50}
\definecolor{cat4color}{HTML}{9C27B0}

\theoremstyle{plain}
\newtheorem{theorem}{Theorem}[section]
\newtheorem{proposition}[theorem]{Proposition}
\newtheorem{lemma}[theorem]{Lemma}

\theoremstyle{definition}
\newtheorem{definition}[theorem]{Definition}

\theoremstyle{remark}

\title{Evaluating Uncertainty Calibration in Probabilistic Time Series Foundation Models}

\author{Thomas Decker$^{1,2,3}$  \qquad \textbf{Volker Tresp}$^{2,3}$ \\
	$^1$Siemens AG \quad $^2$LMU Munich \quad $^3$Munich Center for Machine Learning (MCML) \\
	\texttt{thomas.decker@siemens.com}, \qquad \texttt{volker.tresp@lmu.de}
}

\begin{document}
	
	\maketitle
	
	\begin{abstract}
		Time series foundation models (TSFMs) are increasingly used for probabilistic forecasting across diverse domains, yet the reliability of their uncertainty estimates remains poorly understood. In this work, we study uncertainty calibration in probabilistic TSFMs from a broader methodological and empirical perspective. We introduce a taxonomy of how TSFMs represent predictive uncertainty and develop a fine-grained evaluation framework for calibration in multi-horizon forecasting that distinguishes pooled, horizon-wise, and stronger conditional notions of calibration. Using this framework, we conduct a comprehensive empirical evaluation of popular TSFMs across diverse datasets and show that current models exhibit distinct and systematic patterns of miscalibration that standard metrics often obscure. We further analyze common post-hoc recalibration methods and characterize which forms of calibration they can improve and which tend to persist. Our results provide practical guidance for evaluating and improving uncertainty calibration in TSFMs and highlight important limitations of current practice.
	\end{abstract}

	\section{Introduction}
	\label{sec:introduction}
	
	Time series foundation models (TSFMs) are progressively emerging as a general-purpose approach to forecasting \cite{liang2024foundation}. Pretrained on large and heterogeneous collections of time series, they can often be applied in a zero-shot manner to unseen datasets, frequencies, and domains while remaining competitive with specialized forecasting systems \cite{ansari2024chronos, woo2024moirai, das2024timesfm}. As their adoption grows in high-stakes applications such as healthcare \cite{gu2025vitalsign}, energy systems \cite{meyer2025benchmarking}, and finance \cite{zhu2025fincast}, it becomes increasingly important that they not only produce accurate forecasts but also communicate uncertainty in a trustworthy manner.
	
	In many of these settings, decisions depend not only on the predicted outcome itself but also on how likely adverse outcomes are. This makes the quality of uncertainty estimates an important practical objective beyond predictive accuracy. However, such estimates are useful only if their stated confidence levels can be interpreted reliably. In particular, a prediction interval reported at \(90\%\) confidence for a forecast should contain the realized outcome about \(90\%\) of the time. This notion of reliability is commonly captured by calibration, meaning that predicted probabilities agree with observed frequencies \cite{gneiting2007probabilistic, gruber2022better, kuleshov2018calibrated, song2019distributioncalibration}. A model might therefore achieve strong forecasting accuracy while still providing uncertainty estimates that are misleading in practice.
	
	Although TSFMs increasingly support uncertainty-aware forecasting, calibration in this setting is still only partially understood. Recent work has begun to examine this question empirically \cite{adler2025beyondaccuracy}, providing an important first view of calibration in modern forecasting models. At the same time, calibration is a subtle property in multi-horizon forecasting, where conclusions can depend on how forecasts are represented, how behavior varies across horizon steps, and how results are aggregated across instances. This creates the need for a more systematic evaluation framework that supports comparison across heterogeneous TSFMs and reveals forms of miscalibration that may be hidden by standard aggregate metrics.
	
	This question is also practically important because miscalibration is often addressed after training through post-hoc recalibration methods. Such techniques have been studied extensively in classification \cite{minderer2021revisiting, silva2023classifier} and regression \cite{kuleshov2018calibrated, song2019distributioncalibration, dheur2023largescale} as a lightweight way to improve the reliability of model confidence estimates without significantly retraining the underlying predictor. However, for probabilistic TSFMs, their role remains unclear as it is not well understood yet how effective they can be on individual datasets, and which forms of miscalibration they are able to correct.

\begin{figure}
		\centering
		\includegraphics[width=1\linewidth]{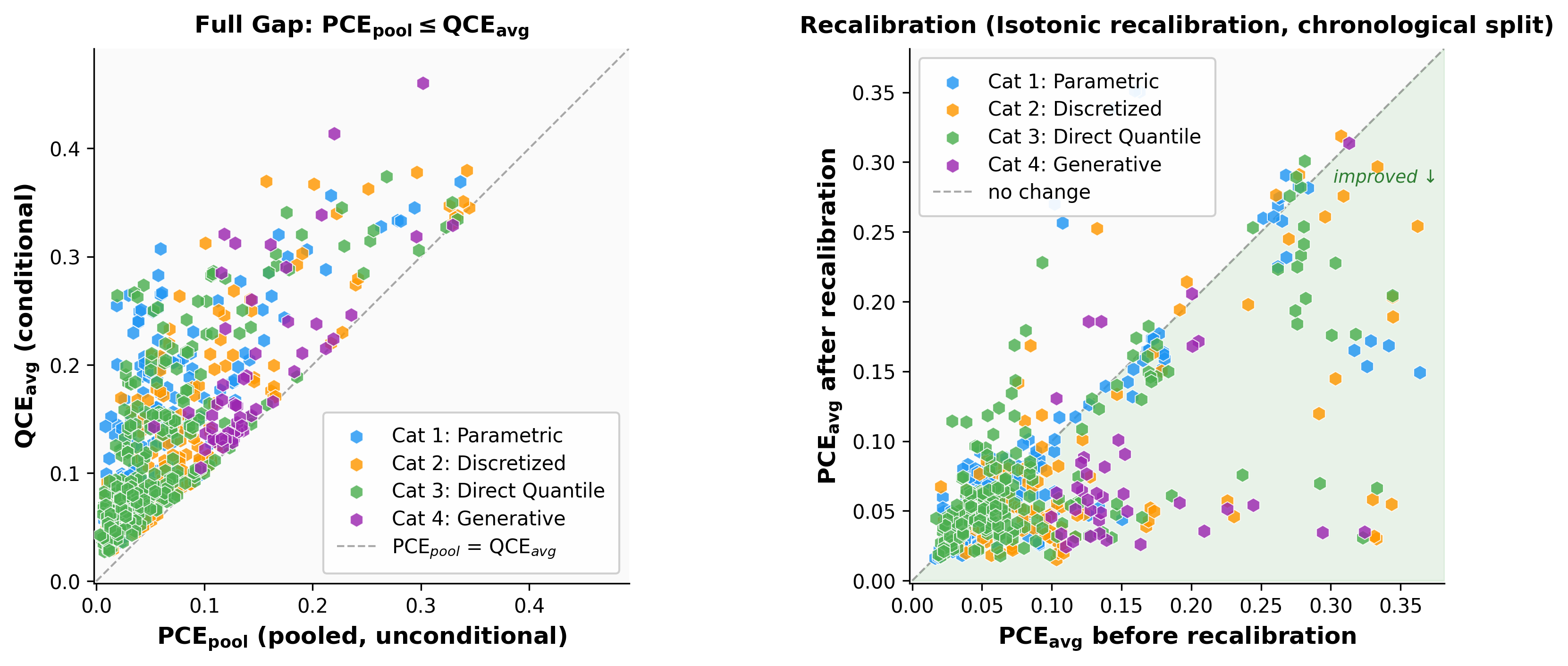}
		\vspace{-0.5cm}
		\caption{Overview of the proposed calibration evaluation framework and main empirical findings. \textbf{Left:} Relationship between \( \mathrm{PCE}_{\mathrm{pool}} \) and \( \mathrm{QCE}_{\mathrm{avg}} \) across model-dataset pairs. All points lie above the diagonal, consistent with the calibration hierarchy \( \mathrm{PCE}_{\mathrm{pool}} \le \mathrm{QCE}_{\mathrm{avg}} \), showing that pooled probabilistic calibration can substantially understate miscalibration once horizon-specific and quantile-conditioned effects are taken into account. \textbf{Right:} Effect of post-hoc isotonic recalibration on \( \mathrm{PCE}_{\mathrm{avg}} \), fitted on the earliest 20\% of the forecast windows and evaluated on the later ones. Points below the diagonal improve (62\% of the model--dataset pairs), and the largest gains occur where the original miscalibration is most severe, indicating that even simple recalibration techniques often improve horizon-wise probabilistic calibration in practice.}
		
		\label{fig:overview}
		\vspace{-0.5cm}
	\end{figure}

	In this work, we present a systematic study of uncertainty calibration in probabilistic TSFMs. We first introduce a taxonomy of uncertainty quantification mechanisms that organizes TSFMs by the predictive object from which their uncertainty estimates are obtained. Building on this common view, we then develop a calibration evaluation framework for multi-horizon forecasting that goes beyond pooled average coverage and includes stronger quantile-conditioned diagnostics. Finally, we use this framework to analyze calibration and post-hoc recalibration across a broad collection of TSFMs and datasets. Figure~\ref{fig:overview} summarizes the framework and the main empirical findings.
	
	Our main contributions are as follows:
	\begin{itemize}[leftmargin=12pt, itemsep=2pt]
		\item We introduce a unifying taxonomy of uncertainty quantification mechanisms in probabilistic TSFMs, organizing models by the predictive object from which their horizon-wise predictive distributions are obtained.
		
		\item We develop a calibration evaluation framework for probabilistic multi-horizon forecasting that applies across these model classes and distinguishes pooled, horizon-wise, and quantile-conditioned notions of calibration.
		
		\item We establish theoretical relationships between the resulting calibration metrics, clarifying how pooled and average coverage can hide structured forms of miscalibration, and use these relationships to define diagnostic gap measures for horizon-specific and value-dependent errors.
		
		\item We provide a broad empirical study of calibration and post-hoc recalibration across diverse TSFMs and datasets, showing that current models exhibit systematic but heterogeneous miscalibration patterns and that standard recalibration methods often improve average calibration substantially while only partially addressing stronger conditional deficiencies.
	\end{itemize}
	
	\section{Background and Related Work}
	\label{sec:background}
	
	\paragraph{Mathematical setup}
	We consider a univariate time series as an ordered sequence of real-valued observations. In forecasting, we observe the most recent $T \in \mathbb{N}$ values, called the \textit{context window} $\mathbf{x} = (x_1, \dots, x_T) \in \mathbb{R}^T$, and the goal is to predict the next $H \in \mathbb{N}$ values, called the \textit{forecast horizon} $\mathbf{y} = (y_1, \dots, y_H) \in \mathbb{R}^H$. Since future values are uncertain, we treat $\mathbf{y}$ as a realization of a random vector $\mathbf{Y} = (Y_1, \dots, Y_H)$ and write $P_{\mathbf{Y}\mid \mathbf{X}=\mathbf{x}}$ for its true conditional distribution given context $\mathbf{x}$, where $\mathbf{x}$ is a realization of the random vector $\mathbf{X} = (X_1, \dots, X_T)$.
	
	For scalar-valued targets, uncertainty is often represented through the cumulative distribution function (CDF) and its inverse, the quantile function. For a random variable $Y$ with realization $y \in \mathbb{R}$, these are defined as
	\begin{align*}
		F_Y(y) &:= P(Y \le y), \qquad
		F_Y^{-1}(\tau) := \inf\{y \in \mathbb{R} : F_Y(y) \ge \tau\},
	\end{align*}
	where $F_Y(y)$ gives the probability that $Y$ does not exceed $y$, and $F_Y^{-1}(\tau)$ gives the threshold below which a fraction $\tau$ of outcomes fall. Quantiles are especially useful in forecasting because they directly induce prediction intervals. At nominal level $1-\alpha$, the interval $[F_Y^{-1}(\alpha/2),\, F_Y^{-1}(1-\alpha/2)]$ will contain the realized outcome with probability $1-\alpha$ under the true distribution if $F_Y$ is continuous.
	
	The goal of probabilistic forecasting is therefore to estimate, for each context \( \mathbf{x} \), the conditional predictive distribution of future values. In this work, we focus on the corresponding horizon-wise marginal predictive CDFs \( \hat{F}_{Y_h \mid \mathbf{X}=\mathbf{x}} \), \( h=1,\dots,H \), as the common probabilistic object that is accessible across the different TSFM families considered below. The mechanism through which these predictive CDFs are represented, however, differs substantially across models.
	
	\paragraph{Uncertainty calibration in forecasting}
	A probabilistic forecast is calibrated if its stated probabilities agree with empirical frequencies \cite{gneiting2007probabilistic, gruber2022better}. In probabilistic regression, a standard notion is probabilistic calibration \cite{kuleshov2018calibrated,dheur2023largescale}, which requires predicted quantiles to attain the advertised coverage rate on average across forecast instances. For a one-step forecast and quantile level $\tau \in (0,1)$, this can be written as
	\begin{align*}
		\frac{1}{N}\sum_{n=1}^{N}
		\mathbf{1}\!\left\{Y_n \le \hat{F}^{-1}_{Y \mid \mathbf{X}=\mathbf{x}_n}(\tau)\right\}
		\approx \tau.
	\end{align*}
	That is, across $N$ forecast instances, the fraction of observed outcomes below the predicted $\tau$-quantile should be close to $\tau$. Averaging the deviation from this identity over a finite set of quantile levels $\mathcal{T} \subset (0,1)$ yields the probabilistic calibration error (PCE) \cite{dheur2023largescale}:
	\begin{align*}
		\mathrm{PCE}
		=
		\frac{1}{|\mathcal{T}|}
		\sum_{\tau \in \mathcal{T}}
		\left|
		\frac{1}{N}\sum_{n=1}^{N}
		\mathbf{1}\!\left\{Y_n \le \hat{F}^{-1}_{Y \mid \mathbf{X}=\mathbf{x}_n}(\tau)\right\}
		- \tau
		\right|.
	\end{align*}
	
	In multi-horizon forecasting, each prediction produces $H$ forecast steps, and calibration can be summarized across horizons in different ways. The authors of \cite{adler2025beyondaccuracy} pool all horizon-instance pairs into a single collection, yielding a global calibration measure across the full test set. This gives a useful aggregate view, but it can mask systematic differences across horizons or other subsets of instances. In particular, a model may appear well calibrated under pooled average coverage while still exhibiting structured miscalibration in specific regions of its predictive distribution.
	
	This limitation is closely related to the distinction between average and conditional notions of calibration \cite{gneiting2007probabilistic,gneiting2023regression}. Probabilistic calibration only requires coverage to hold on average across all instances, so overconfidence in some regions can be offset by underconfidence in others. Stronger conditional notions ask whether predicted probabilities remain reliable given the model's own forecasted level, rather than only after averaging over all predictions. Our work builds on this perspective and studies calibration for TSFMs at a finer level than pooled average coverage alone.

	\paragraph{Post-hoc recalibration.}
	When predictive uncertainty is miscalibrated, a practical question is whether it can be corrected without retraining the forecasting model. Post-hoc recalibration addresses this by learning a lightweight correction on held-out data and applying it after training. Such methods are particularly attractive for TSFMs, which are large, pretrained, and often used in zero-shot mode. Popular approaches include isotonic recalibration \cite{kuleshov2018calibrated}, Platt-style scaling \cite{platt1999probabilistic}, distribution-based recalibration methods \cite{song2019distributioncalibration}, and KDE-based methods \cite{dheur2023largescale}. While these approaches have shown strong empirical performance in regression settings \cite{dheur2023largescale}, their applicability and limitations for TSFMs remain largely unexplored.
	
	\section{A Unifying Taxonomy of Probabilistic TSFMs}
	\label{sec:taxonomy}
	
	Probabilistic TSFMs differ not only in architecture and pretraining data, but also in the predictive object through which they represent forecast uncertainty. To compare calibration across such heterogeneous models, we begin by introducing a unifying perspective on these approaches. What they have in common is that, for each forecast horizon step \(h=1,\dots,H\), every model ultimately provides an estimate of the marginal predictive CDF \( \hat{F}_{Y_h \mid \mathbf{X}=\mathbf{x}} \). The key difference lies in the intermediate representation from which this CDF is obtained and in the training objective used to learn it. Based on this perspective, we distinguish four conceptually different categories.

	\paragraph{Category 1: Parametric density models.}
	In this category, TSFMs represent predictive uncertainty through an explicit parametric family. For each horizon step \(h\), the model predicts parameters \( \hat{\theta}_h \) of a conditional density
	$\hat{p}_{Y_h \mid \mathbf{X}=\mathbf{x}}(y)=p(y;\hat{\theta}_h)$.
	The predictive CDF is then accessible by integration,
	\begin{align*}
		\hat{F}_{Y_h \mid \mathbf{X}=\mathbf{x}}(y)
		=
		\int_{-\infty}^{y} p(z;\hat{\theta}_h)\,dz.
	\end{align*}
	For autoregressive heads, later-horizon marginals are obtained by sampling and need not remain in the parametric family. Training in this category is naturally based on negative log-likelihood. At inference time, quantiles and prediction intervals may be available analytically for simple families, but are often obtained numerically or through sampling in practice. This mechanism is, for instance, used by \textbf{Lag-Llama} \cite{rasul2023lagllama}, which employs a Student-\(t\) output head, by \textbf{MOIRAI 1.x} and \textbf{MOIRAI-MoE} \cite{woo2024moirai,liu2024moiraimoe}, which use mixture heads, or by \textbf{Toto 1.0} \cite{cohen2025toto}, which uses a Student-\(t\) mixture head.
	
	\paragraph{Category 2: Discretized distribution models.}
	In this category, TSFMs represent predictive uncertainty over a finite discretization of the real value axis. Let $b_0 < b_1 < \cdots < b_K$
	denote bin boundaries in $\mathbb{R}$. For each horizon step \(h\), a model in this category predicts probabilities
	\begin{align*}
		\hat{p}_{h,k}
		=
		\hat{P}\bigl(Y_h \in (b_{k-1},b_k] \mid \mathbf{X}=\mathbf{x}\bigr),
		\qquad k=1,\dots,K.
	\end{align*}
	This is the mechanism used by \textbf{Chronos-T5} \cite{ansari2024chronos}, which quantizes values into tokens and predicts token probabilities autoregressively, and by \textbf{TabPFN-TS} \cite{hoo2025tabpfnts}, whose bar distribution assigns predictive mass to fixed intervals.
	For token-based models the predictive CDF is then formed by cumulative summation over the discretized support, yielding a stepwise approximation:
	\begin{align*}
		\hat{F}_{Y_h \mid \mathbf{X}=\mathbf{x}}(y)
		=
		\sum_{k:\, b_k \le y} \hat{p}_{h,k}.
	\end{align*}
	For bar distributions, the mass is assumed uniform within each bin, so the CDF is not a staircase but increases linearly within each bin. Training is typically carried out with cross-entropy, equivalently a discretized log-likelihood under the induced representation. At inference time, quantiles are obtained by direct inversion of the CDF, whereas autoregressive token-based variants generate forecast values by sampling from the per-step categorical distribution.
	
	\paragraph{Category 3: Direct quantile forecasters.}
	In this category, TSFMs directly predict values of the inverse CDF at a prescribed finite set of probability levels \( \tau_1,\dots,\tau_M \in (0,1) \). For each horizon step \(h\), the model outputs
	\begin{align*}
		\hat{F}^{-1}_{Y_h \mid \mathbf{X}=\mathbf{x}}(\tau_1),\dots,
		\hat{F}^{-1}_{Y_h \mid \mathbf{X}=\mathbf{x}}(\tau_M).
	\end{align*}
	The model is therefore trained to estimate selected points of the inverse CDF directly rather than a full density or CDF over the outcome space. The standard objective is the pinball loss or weighted variants \cite{chung2021beyond}. At inference time, uncertainty is typically available only at the chosen quantile levels, so its resolution is limited by the quantile grid used during training, although some models additionally offer a continuous quantile head. This is a popular approach used by many recent models such as \textbf{Chronos-Bolt} \cite{ansari2024chronosbolt}, \textbf{Chronos-2} \cite{ansari2025chronos2}, \textbf{MOIRAI 2.0} \cite{aksu2025moirai2}, \textbf{TiRex} \cite{auer2025tirex}, \textbf{TimesFM} \cite{das2024timesfm}, \textbf{KAIROS} \cite{feng2025kairos}, \textbf{FlowState} \cite{graf2025flowstate}, and \textbf{YINGLONG} \cite{wang2025yinglong}.
	
	\paragraph{Category 4: Implicit generative process models.}
	The defining feature of this category is that predictive uncertainty is represented implicitly and is only accessible through sampling. Corresponding TSFMs learn a generative process for entire forecast trajectories $\hat{\mathbf{y}}^{(1)},\dots,\hat{\mathbf{y}}^{(S)},
	\quad
	\hat{\mathbf{y}}^{(s)} \in \mathbb{R}^H$
	which result as samples from an implicit joint predictive distribution over the full horizon. The horizon-wise predictive CDF is then estimated empirically by:
	\begin{align*}
		\hat{F}_{Y_h \mid \mathbf{X}=\mathbf{x}}(y)
		=
		\frac{1}{S}\sum_{s=1}^{S}
		\mathbf{1}\!\left\{
		\hat{y}^{(s)}_h \le y
		\right\}.
	\end{align*}
	This mechanism is for instance used by \textbf{Sundial} \cite{liu2025sundial}, which learns a conditional generative transport process via flow matching \cite{lipman2023flow}, and by \textbf{TimeDiT} \cite{cao2024timedit}, which generates trajectories using a diffusion-based denoising objective \cite{ho2020denoising}. In both cases, quantiles and prediction intervals need to be computed from the empirical distribution of sampled paths rather than from an explicit probabilistic output.
	
	\medskip
	
	This taxonomy separates TSFMs by the intermediate predictive object from which the horizon-wise CDF is obtained, as also summarized in Table~\ref{tab:taxonomy}. This distinction is central for calibration analysis as it clarifies which aspects of the predictive distribution are directly available for evaluation and which calibration diagnostics and post-hoc correction methods naturally apply.
	
	\begin{table}[t]
		\centering
		\small
		\setlength{\tabcolsep}{5pt}
		\renewcommand{\arraystretch}{1.35}
		\begin{tabular}{@{} c >{\bfseries\raggedright\arraybackslash}p{1.9cm} p{5.3cm} p{2.0cm} p{3.6cm} @{}}\toprule
			\textbf{\#} & \textbf{Category} & \textbf{How $\hat{F}_{Y_h \mid \mathbf{X}=\mathbf{x}}$ is obtained} & \textbf{Training loss} & \textbf{Representative models} \\
			\midrule
			
			\rowcolor{cat1color!12}
			1 & Parametric Density
			& The model predicts parameters $\hat{\theta}_h$ of a conditional density family $p(y;\hat{\theta}_h)$; the predictive CDF may be obtained analytically or via sampling in practice.
			& Negative log-likelihood
			& Lag-Llama \cite{rasul2023lagllama}, MOIRAI 1.x \cite{woo2024moirai}, MOIRAI-MoE \cite{liu2024moiraimoe}, Toto 1.0 \cite{cohen2025toto} \\
			
			\rowcolor{cat2color!12}
			2 & Discretized Distribution
			& The model predicts discrete probabilities $\hat{p}_{h,k}$ over predefined value intervals; the predictive CDF is obtained by cumulative mass over the discretized support, interpolated within bins for bar distributions.
			& Cross-entropy / discretized log-likelihood
			& Chronos-T5 \cite{ansari2024chronos}, TabPFN-TS \cite{hoo2025tabpfnts} \\
			
			\rowcolor{cat3color!12}
			3 & Direct Quantile
			& The model predicts $\hat{F}^{-1}(\tau)$ at finitely many levels $\tau_1,\dots,\tau_M$; uncertainty is therefore typically available only at the resolution of the chosen quantile grid.
			& Pinball loss / weighted quantile loss
			& Chronos-Bolt \cite{ansari2024chronosbolt}, Chronos-2 \cite{ansari2025chronos2}, MOIRAI 2.0 \cite{aksu2025moirai2}, TiRex \cite{auer2025tirex}, TimesFM \cite{das2024timesfm}, KAIROS \cite{feng2025kairos}, FlowState \cite{graf2025flowstate}, YINGLONG \cite{wang2025yinglong} \\
			
			\rowcolor{cat4color!12}
			4 & Implicit Generative Process
			& The model generates forecast trajectories from an implicit joint predictive law; marginal predictive CDFs are estimated empirically from Monte Carlo samples.
			& Flow matching / diffusion denoising
			& Sundial \cite{liu2025sundial}, TimeDiT \cite{cao2024timedit} \\
			
			\bottomrule
		\end{tabular}
		\caption{Taxonomy of uncertainty quantification approaches in TSFMs, organized by the intermediate predictive mechanism learned by the model (density parameters, discretized probability mass, quantile values, or samples), how the predictive CDF is obtained, and the corresponding training objective.}
		\label{tab:taxonomy}
		\vspace{-0.5cm}
	\end{table}

	\section{A Calibration Evaluation Framework for Probabilistic TSFMs}
	\label{sec:calibration_framework}
	
	We now introduce a calibration evaluation framework tailored to probabilistic TSFMs. The goal is to evaluate uncertainty estimates in a way that is both comparable across the heterogeneous output types in Section~\ref{sec:taxonomy} and sufficiently fine-grained for multi-horizon forecasting. To do so, we first refine probabilistic calibration assessment to account for the multi-horizon setting, and then move to a stronger conditional notion of calibration that captures whether uncertainty estimates are also trustworthy at the level of individual predictions. Throughout, we fix a finite set of quantile levels \( \mathcal{T} = \{\tau_1,\dots,\tau_Q\} \subset (0,1) \) and write \( \hat q_{h,\tau_q} := \hat{F}^{-1}_{Y_h \mid \mathbf{X}}(\tau_q) \) for the grid levels. 
	
	\paragraph{Multi-horizon probabilistic calibration.}
	
	A natural extension of the probabilistic calibration error from Section~\ref{sec:background} is to pool all horizon-instance pairs and compare empirical frequencies with nominal quantile levels over the full test set. Following \cite{adler2025beyondaccuracy}, we define
	\begin{align*}
		\mathrm{PCE}_{\mathrm{pool}}
		=
		\frac{1}{Q}\sum_{q=1}^{Q}
		\left|
		\frac{1}{H}\sum_{h=1}^{H}
		P\!\left(Y_h \le \hat q_{h,\tau_q}\right)
		- \tau_q
		\right|.
		\label{eq:pce_pool}
	\end{align*}
	This provides a concise global summary, but it can hide systematic differences across forecast steps because over- and undercoverage at different horizons may cancel after pooling. To avoid this, we evaluate calibration separately at each horizon and aggregate the resulting errors by their mean and maximum. For $h=1, \dots, H$ we define:
	\begin{align*}
		\mathrm{PCE}_h
		=
		\frac{1}{Q}\sum_{q=1}^{Q}
		\left|
		P\!\left(Y_h \le \hat q_{h,\tau_q}\right)
		- \tau_q
		\right|
	\end{align*}
	such that  \( \mathrm{PCE}_{\mathrm{avg}} = \frac{1}{H}\sum_{h=1}^{H}\mathrm{PCE}_h \) summarizes the typical error across the horizon, while \( \mathrm{PCE}_{\max}= \max_{h=1,\dots,H}\mathrm{PCE}_h \) identifies the worst forecast step.
	
	\paragraph{Quantile-conditioned calibration.}
	
	Even when probabilistic calibration error is small, a model may still be unreliable in a more structured way. Its predictive intervals may be too narrow for some predicted quantile values and too wide for others, with these effects averaging out overall. To capture such value-dependent distortions, we consider the stronger notion of  quantile-conditioned calibration \cite{gneiting2023regression}, which requires calibration to hold conditional on the model's own predicted quantile value.
	
	\begin{definition}[Quantile-conditioned calibration]
		\label{def:conditional_quantile_calibration}
		A predictive CDF \( \hat{F} \) for a real-valued outcome \( Y \) is quantile-conditioned calibrated if
		\begin{align*}
			P\!\left(Y \le \hat{F}^{-1}(\tau) \;\middle|\; \hat{F}^{-1}(\tau)\right) = \tau
			\qquad \text{for all } \tau \in (0,1), \text{ almost surely.}
		\end{align*}
	\end{definition}
	In words, among all instances for which the model predicts similar values for the \( \tau \)-quantile, the realized outcome should fall below that value a fraction \( \tau \) of the time. For continuous predictive CDFs, this is the conditional exceedance probability (CEP) calibration discussed in \cite{gneiting2023regression}. 
	This notion is stronger than average probabilistic calibration because it conditions on the model output itself rather than averaging over all predictions. At the same time, it remains practically approachable through nonparametric estimation (see Appendix~\ref{sec:calib_computation}), which makes it useful as a diagnostic even though it is more challenging to estimate than average coverage. We then define
	\begin{align*}
		\mathrm{QCE}_h
		=
		\frac{1}{Q}\sum_{q=1}^{Q}
		\mathbb{E}\!\left[
		\left|
		P\!\left(Y_h \le \hat q_{h,\tau_q} \;\middle|\; \hat q_{h,\tau_q}\right)
		- \tau_q
		\right|
		\right],
		\qquad h=1,\dots,H,
	\end{align*}
	and aggregate again via
	\begin{align*}
		\mathrm{QCE}_{\mathrm{avg}} = \frac{1}{H}\sum_{h=1}^{H}\mathrm{QCE}_h,
		\qquad
		\mathrm{QCE}_{\max} = \max_{h=1,\dots,H}\mathrm{QCE}_h.
	\end{align*}
	Thus, \( \mathrm{QCE}_{\mathrm{avg}} \) and \( \mathrm{QCE}_{\max} \) are horizon-aware counterparts of PCE based on a stronger calibration requirement that can still be estimated in practice. This is particularly useful for TSFMs because models with similar values of \( \mathrm{PCE}_{\mathrm{avg}} \) can still differ substantially in how reliable their uncertainty estimates are across different kinds of predictions. We estimate it by isotonic regression, whose population target lower-bounds \( \mathrm{QCE}_h \), with equality for non-decreasing conditional coverage (see Appendix~\ref{sec:calib_computation}).
	
	\paragraph{A hierarchy of calibration errors.}
	
	The metrics introduced above form a natural ordering which can be expressed by the following proposition:
	
	\begin{proposition}
		\label{thm:hierarchy}
		For any predictive model and any finite quantile grid \( \mathcal{T} \),
		\begin{align*}
			\mathrm{PCE}_{\mathrm{pool}}
			\;\le\;
			\mathrm{PCE}_{\mathrm{avg}}
			\;\le\;
			\mathrm{QCE}_{\mathrm{avg}}
			\;\le\;
			\mathrm{QCE}_{\mathrm{max}}.
		\end{align*}
	\end{proposition}
	
	This proposition clarifies the role of the different metrics. \( \mathrm{PCE}_{\mathrm{pool}} \) is the weakest summary because it aggregates over all horizons and instances at once. \( \mathrm{PCE}_{\mathrm{avg}} \) is stricter because it evaluates each horizon separately before aggregation. \( \mathrm{QCE}_{\mathrm{avg}} \) is stricter still because it assesses whether calibration remains valid conditional on the model's own predicted quantile values. Finally, \( \mathrm{QCE}_{\max} \) upper-bounds all of them and highlights the worst-calibrated forecast horizon under the strongest notion considered here.

	\begin{definition}[Horizon gap and conditioning gap]
		\label{def:gaps}
		We define the \emph{horizon gap} \( \Delta_H \) and \emph{conditioning gap} \( \Delta_C \) as additional diagnostics by
		\begin{align*}
			\Delta_H := \mathrm{PCE}_{\mathrm{avg}} - \mathrm{PCE}_{\mathrm{pool}} \ge 0,
			\qquad
			\Delta_C := \mathrm{QCE}_{\mathrm{avg}} - \mathrm{PCE}_{\mathrm{avg}} \ge 0.
		\end{align*}
	\end{definition}
	
	The horizon gap \( \Delta_H \) measures how much calibration error is hidden by pooling across horizons. A large value indicates that reliability differs substantially across forecast steps, even if the pooled summary appears acceptable. The conditioning gap \( \Delta_C \) measures how much additional error becomes visible once one conditions on the model's own predicted quantile values. A large conditioning gap therefore points to structured value-dependent miscalibration that is invisible under average coverage alone. Together, these diagnostics help separate horizon-specific effects from deeper conditional unreliability and will also be useful when interpreting post-hoc recalibration results. Since finite-sample estimates of both gaps are positive even when no such structure is present, we accompany them with significance tests (see Appendix~\ref{app:significance}).
	
	\paragraph{Implications for post-hoc recalibration.}
	
	The proposed evaluation framework can be used not only to assess how well TSFMs are calibrated out of the box, but also to analyze which types of miscalibration can be improved by post-hoc correction.  Quantile relabeling methods \cite{dheur2023largescale} are particularly appealing for probabilistic TSFMs as t hese methods preserve the support and ordering of predicted outcomes and only redistribute probability mass to improve calibration. This makes them lightweight and leaves the pretrained model itself unchanged. Formally, such methods estimate a monotone map $g:[0,1]\to[0,1]$ from a small held-out calibration set and redefine
	\begin{align*}
		\hat{F}^{-1,\mathrm{g}}_{Y_h \mid \mathbf{X}}(\tau) := \hat{F}^{-1}_{Y_h \mid \mathbf{X}}\!\left(g(\tau)\right).
	\end{align*}
	For models that output quantiles only on a finite grid, $\hat{F}^{-1}_{Y_h \mid \mathbf{X}}$ is interpolated between the grid levels.
	To understand how such recalibration interacts with our diagnostic framework, we characterize which forms of miscalibration can and cannot be addressed by a single global relabeling map.
	\begin{theorem}
		\label{thm:global_relabelling}
		Let \( g:[0,1]\to[0,1] \) be any relabeling map, and let
		\( \mathrm{PCE}_{\mathrm{pool}}^g \), \( \mathrm{PCE}_{\mathrm{avg}}^g \), and
		\( \mathrm{QCE}_{\mathrm{avg}}^g \) denote the corresponding calibration errors
		after recalibration with \( g \). Then the following hold:
		\begin{enumerate}
			\item[(i)] If the pooled coverage \( \bar p(t):=\frac1H\sum_{h=1}^H P\bigl(Y_h\le \hat F^{-1}_{Y_h\mid\mathbf X}(t)\bigr) \) is continuous in \(t\), there exists a non-decreasing map \( g^\star \) with
			\( \mathrm{PCE}_{\mathrm{pool}}^{g^\star}=0 \).
			
			\item[(ii)] For every \( g \), \( \mathrm{PCE}_{\mathrm{avg}}^{g} \ge \Delta_H/2 \), where \( \Delta_H \) is the horizon gap of the original forecast. In particular, if the original forecast exhibits a non-zero horizon gap
			\( \Delta_H > 0 \), then \( \mathrm{PCE}_{\mathrm{avg}}^{g} > 0 \) for every
			\( g \), including any \( g \) with
			\( \mathrm{PCE}_{\mathrm{pool}}^{g}=0 \).
			
			\item[(iii)] For every \( \varepsilon \in (0,1/2) \), there exists a forecasting
			setup with continuous \( \bar p \) in which every \( g \) with
			\( \mathrm{PCE}_{\mathrm{pool}}^{g}=0 \) satisfies
			\( \mathrm{QCE}_{\mathrm{avg}}^{g}-\mathrm{QCE}_{\mathrm{avg}}>\varepsilon \).
		\end{enumerate}
	\end{theorem}
	
	This theorem highlights key limitations of global relabeling-based recalibration. In principle, relabeling is able to eliminate the pooled probabilistic calibration error entirely, a map that isotonic recalibration approximates by estimating the pooled coverage curve. However, this is the weakest notion in our hierarchy. If miscalibration differs across horizons, then no global relabeling can remove horizon-wise errors simultaneously. At each quantile level, the map moves the coverage of all horizons in the same direction, so it can reduce over-coverage at some horizons only by increasing under-coverage at others. Part~(ii) shows that half of the horizon gap is a floor for \( \mathrm{PCE}_{\mathrm{avg}} \) that no global relabeling can go below. Moreover, success under pooled coverage does not imply improvement under the stronger quantile-conditioned notion. Part~(iii) shows that there are setups in which every map that removes the pooled error increases \( \mathrm{QCE}_{\mathrm{avg}} \) by almost \(1/2\). This additionally motivates the proposed hierarchy for evaluating recalibration of probabilistic TSFMs as pooled summaries alone may overstate the extent to which uncertainty estimates have truly been improved. All proofs, together with the intuition behind them, are given in Appendix~\ref{app:proofs}.
	
	\section{Experiments}
	\label{sec:experiments}
	\paragraph{Setup.}
	
	We now apply the proposed calibration framework to a comprehensive empirical benchmark. We evaluate 14 zero-shot probabilistic TSFMs spanning all four uncertainty quantification categories from Section~\ref{sec:taxonomy}: five parametric density models, three discretized distribution models, five direct quantile models, and one implicit generative model. Here, zero-shot means that no model is finetuned on the evaluated datasets. Each model is tested on the short-term forecasting task of 55 dataset configurations from the GIFT-Eval benchmark \cite{aksu2024gift} covering diverse domains and sampling frequencies. The recalibration experiments additionally require at least 50 forecast windows per dataset, which leaves 42 configurations. To ensure comparability across model classes, all calibration metrics are computed on the common quantile grid \( \mathcal{T}=\{0.1,0.2,\dots,0.9\} \). For models that do not directly expose these quantiles, they are extracted from the corresponding predictive distribution or empirical samples. Quantile-conditioned calibration errors are estimated nonparametrically by isotonic regression of the coverage indicators on the predicted quantile values, and we test the significance of both diagnostic gaps for every model--dataset pair with at least two series. Full implementation details are provided in Appendix~\ref{app:experimental_details} and accompanying code is available at \texttt{https://github.com/thomdeck/tsfm-calibration}.
	
	\paragraph{Fine-grained calibration analysis.}
	
	Table~\ref{tab:scorecard} reports calibration results aggregated across datasets for all evaluated TSFMs. We highlight three observations. First, \( \mathrm{PCE}_{\mathrm{avg}} \) is consistently larger than \( \mathrm{PCE}_{\mathrm{pool}} \) (in 98\% of the model--dataset pairs, and by 40\% on average), showing that pooled coverage systematically understates calibration error by averaging over forecast steps with different reliability profiles. Second, the further increase to \( \mathrm{QCE}_{\mathrm{avg}} \) is substantial across all model families, indicating that even horizon-wise average coverage misses structured value-dependent errors that only become visible under the stronger quantile-conditioned criterion. The gap diagnostics make this explicit. \( \Delta_H \) captures uneven calibration across the horizon, whereas \( \Delta_C \) captures residual conditional distortions beyond average coverage. Both gaps are significant for most model--dataset pairs with at least 50 forecast windows (\(\Delta_H\): 86\%, \(\Delta_C\): 69\%, see also Appendix~\ref{app:significance}). Third, parametric and direct quantile models tend to display the strongest calibration profile across the considered metrics, although the differences remain dataset-dependent. Chronos-T5 and Sundial combine large pooled errors with small corrected conditioning gaps, so their miscalibration appears to be mostly marginal. Sundial's nominal 80\% interval, for instance, covers only 40\% of the outcomes.
	
	Figure~\ref{fig:overview}(left) complements this aggregate view by showing \( \mathrm{PCE}_{\mathrm{pool}} \) against \( \mathrm{QCE}_{\mathrm{avg}} \) for individual model-dataset pairs. Consistent with the calibration hierarchy, all points lie above the diagonal. This gap varies considerably across datasets, indicating that the extent of hidden miscalibration is highly setting-dependent. Further metric pairs are shown in Appendix~\ref{app:additional_results}.

	\begin{table}[t]
		\centering
		\small
		\setlength{\tabcolsep}{4pt}
		\renewcommand{\arraystretch}{1.2}
		\resizebox{\textwidth}{!}{%
			\begin{tabular}{@{} c l r r r r r r r r r r @{}} \toprule
				& & & & & & & & \multicolumn{2}{c}{\(\mathbf{\Delta_C}\)} & \multicolumn{2}{c}{\textbf{\% significant}} \\
				\cmidrule(lr){9-10} \cmidrule(l){11-12}
				\textbf{Cat.} & \textbf{Model} & \( \mathbf{PCE}_{\mathbf{pool}} \) & \( \mathbf{PCE}_{\mathbf{avg}} \) & \( \mathbf{PCE}_{\mathbf{max}} \) & \( \mathbf{QCE}_{\mathbf{avg}} \) & \( \mathbf{QCE}_{\mathbf{max}} \) & \( \mathbf{\Delta_H} \) & \textbf{raw} & \textbf{corr.} & \( \mathbf{\Delta_H} \) & \( \mathbf{\Delta_C} \) \\
				\midrule
				1 & \cellcolor{cat1color!12}\textbf{Moirai-Base} & 0.0536 & 0.0829 & 0.1579 & 0.1215 & 0.1862 & 0.0293 & 0.0386 & 0.0199 & 76\% & 76\% \\
				1 & \cellcolor{cat1color!12}\textbf{Moirai-Large} & 0.0484 & 0.0776 & 0.1581 & 0.1225 & 0.1906 & 0.0292 & 0.0448 & 0.0249 & 80\% & 82\% \\
				1 & \cellcolor{cat1color!12}\textbf{Moirai-MoE-Base} & 0.0533 & 0.0826 & 0.1537 & 0.1204 & 0.1878 & 0.0293 & 0.0379 & 0.0194 & 84\% & 76\% \\
				1 & \cellcolor{cat1color!12}\textbf{Moirai-MoE-Small} & 0.0521 & 0.0837 & 0.1578 & 0.1234 & 0.1908 & 0.0317 & 0.0397 & 0.0215 & 88\% & 76\% \\
				1 & \cellcolor{cat1color!12}\textbf{Moirai-Small} & 0.0530 & 0.0855 & 0.1607 & 0.1283 & 0.1982 & 0.0326 & 0.0428 & 0.0238 & 86\% & 73\% \\
				\midrule
				2 & \cellcolor{cat2color!12}\textbf{Chronos-T5-Base} & 0.1024 & 0.1240 & 0.1939 & 0.1491 & 0.2159 & 0.0216 & 0.0251 & 0.0075 & 84\% & 49\% \\
				2 & \cellcolor{cat2color!12}\textbf{Chronos-T5-Small} & 0.1035 & 0.1253 & 0.1960 & 0.1524 & 0.2183 & 0.0218 & 0.0271 & 0.0092 & 80\% & 53\% \\
				2 & \cellcolor{cat2color!12}\textbf{TabPFN-TS} & 0.0629 & 0.0872 & 0.1537 & 0.1198 & 0.1797 & 0.0243 & 0.0325 & 0.0149 & 78\% & 57\% \\
				\midrule
				3 & \cellcolor{cat3color!12}\textbf{Chronos-2} & 0.0641 & 0.0919 & 0.1653 & 0.1229 & 0.1881 & 0.0278 & 0.0310 & 0.0115 & 78\% & 63\% \\
				3 & \cellcolor{cat3color!12}\textbf{Chronos-Bolt-Base} & 0.0505 & 0.0804 & 0.1494 & 0.1202 & 0.1838 & 0.0299 & 0.0398 & 0.0180 & 88\% & 69\% \\
				3 & \cellcolor{cat3color!12}\textbf{Chronos-Bolt-Small} & 0.0523 & 0.0833 & 0.1569 & 0.1225 & 0.1890 & 0.0310 & 0.0393 & 0.0185 & 84\% & 67\% \\
				3 & \cellcolor{cat3color!12}\textbf{TiRex} & 0.0502 & 0.0796 & 0.1485 & 0.1161 & 0.1798 & 0.0294 & 0.0365 & 0.0174 & 80\% & 69\% \\
				3 & \cellcolor{cat3color!12}\textbf{TimesFM-2.5-200M} & 0.0743 & 0.1011 & 0.1711 & 0.1254 & 0.1900 & 0.0268 & 0.0243 & 0.0058 & 86\% & 45\% \\
				\midrule
				4 & \cellcolor{cat4color!12}\textbf{Sundial-Base} & 0.1467 & 0.1678 & 0.2413 & 0.1958 & 0.2629 & 0.0211 & 0.0280 & 0.0088 & 82\% & 49\% \\
				\bottomrule
			\end{tabular}
		}
		\caption{Calibration metrics across probabilistic TSFMs, averaged over the 55 dataset configurations. Lower is better for all calibration errors and diagnostics. \(\Delta_C\) corr.\ subtracts the part of the conditioning gap expected from estimation noise alone, and the last two columns give the share of model--dataset pairs whose gap is significant at level 0.05. Both use the 49 configurations with at least two series, for which the resampling tests are defined (Appendix~\ref{app:significance}). The category label corresponds to the uncertainty quantification taxonomy in Table~\ref{tab:taxonomy}.}
		\label{tab:scorecard}
		\vspace{-0.2cm}
	\end{table}
	
\begin{figure}[t]
		\centering
		\includegraphics[width=0.99\linewidth]{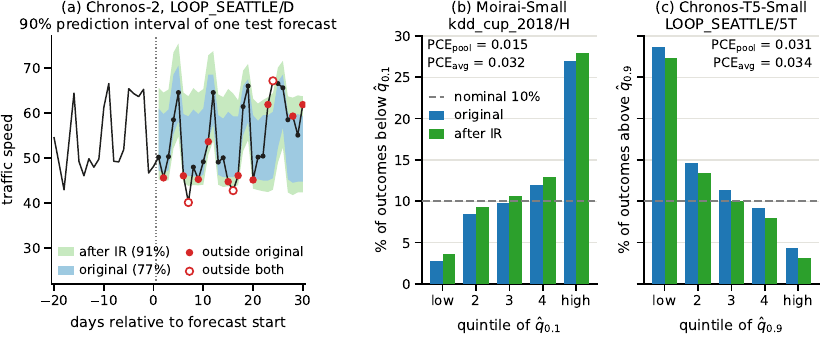}
		
		\caption{Concrete effect of isotonic recalibration (IR). \textbf{(a)} 90\% prediction interval of one forecast before (blue) and after (green) IR; over all test windows, relabeling improves the coverage from 77\% to 91\%. \textbf{(b, c)} Share of outcomes beyond a predicted quantile and differences per quintile of its predicted value. Despite low PCE scores, the actual coverage varies substantially across value ranges.}
		\label{fig:case_study}
		\vspace{-0.2cm}
	\end{figure}

	\paragraph{Why the quantile-conditioned view matters.}
	Figure~\ref{fig:case_study} illustrates what global recalibration can and cannot correct. Marginal miscalibration can matter greatly for individual forecasts: the 90\% intervals of Chronos-2 on \texttt{LOOP\_SEATTLE/D} cover only 77\% of the later outcomes, and isotonic recalibration on the earliest windows raises this to 91\%. Panels (b) and (c) show two pairs that look well calibrated by standard metrics, but grouping the forecasts by their predicted value tells a different story. When Moirai-Small predicts pollution peaks on \texttt{kdd\_cup\_2018/H}, 27\% of the outcomes fall below the predicted 10\% quantile. When Chronos-T5-Small predicts congestion on \texttt{LOOP\_SEATTLE/5T}, 29\% of the speeds exceed the predicted 90\% quantile instead of 10\%. These errors have opposite signs for low and high predictions and cancel in marginal coverage, but not in the conditioning gap (\(\Delta_C=0.039\) and \(0.044\), both significant). Because one relabeling map applies to all forecasts, recalibration shifts all quintiles alike, and the value-dependent pattern persists (\(\Delta_C\): 0.041\(\,\to\,\)0.038 and 0.040\(\,\to\,\)0.038 on the test windows). Removing it would require recalibration that conditions on the forecast itself. Appendix~\ref{app:case_studies} confirms this for further models and datasets.
	
	\paragraph{Post-hoc recalibration effectiveness.}
	We next study how effectively standard post-hoc methods can correct different forms of miscalibration. We evaluate four widely used recalibration methods: isotonic recalibration (IR) \cite{kuleshov2018calibrated}, Platt scaling (PS) \cite{platt1999probabilistic}, histogram binning (HB) \cite{zadrozny2002transforming}, and KDE-based calibration (KDE) \cite{dheur2023largescale}. To reflect a deployment-relevant scenario, we use a chronological split per dataset such that the recalibrator is fitted on the earliest 20\% of the forecast windows of every series and evaluated on the remaining, later windows. For datasets consisting only of single-window series, 20\% of the series are held out for calibration instead (Appendix~\ref{sec:recalibration}). Table~\ref{tab:recalibration} reports the resulting relative changes of the calibration errors, where negative values are improvements. Histogram binning coincides with IR at the fitted quantile levels and is deferred to Appendix~\ref{app:relabeling}.
	
	Across models, all methods improve calibration to a broadly similar extent. As suggested by the theory, the gains are strongest for \( \mathrm{PCE}_{\mathrm{pool}} \) (37\% for IR), which is the calibration notion most directly targeted by global relabeling, smaller for \( \mathrm{PCE}_{\mathrm{avg}} \) (19\%), and smaller still for \( \mathrm{QCE}_{\mathrm{avg}} \) (15\%), while the horizon gap even increases by 24\% and the conditioning gap is essentially unchanged (see Appendix~\ref{app:additional_recal}). This supports the view that post-hoc relabeling is useful for correcting average probabilistic misalignment, but less able to remove deeper value-conditional distortions. Consistently, the models whose miscalibration is mostly marginal benefit most. Chronos-T5, Sundial, and TimesFM lose 51--62\% of their pooled error, compared with 9--24\% for the Moirai and Chronos-Bolt models.
	
	Figure~\ref{fig:overview}(right) shows that \( \mathrm{PCE}_{\mathrm{avg}} \) improves for 62\% of the model--dataset pairs, with the largest gains where the original miscalibration is most severe. Overall, the results indicate that post-hoc recalibration is a practical and data-efficient way to improve the reliability of TSFM uncertainty estimates, but it should be evaluated with more than pooled coverage alone to faithfully judge its effectiveness. 
	
	\begin{table*}[t]
		\centering
		\small
		\setlength{\tabcolsep}{4.5pt}
		\renewcommand{\arraystretch}{1.15}
		\resizebox{\textwidth}{!}{%
			\begin{tabular}{@{} l rrr rrr rrr @{}} \toprule
				& \multicolumn{3}{c}{\textbf{\( \mathbf{PCE}_{\mathbf{pool}} \)}}
				& \multicolumn{3}{c}{\textbf{\( \mathbf{PCE}_{\mathbf{avg}} \)}}
				& \multicolumn{3}{c}{\textbf{\( \mathbf{QCE}_{\mathbf{avg}} \)}} \\
				\cmidrule(lr){2-4} \cmidrule(lr){5-7} \cmidrule(l){8-10}
				\textbf{Model} & \textbf{IR} & \textbf{KDE} & \textbf{PS} & \textbf{IR} & \textbf{KDE} & \textbf{PS} & \textbf{IR} & \textbf{KDE} & \textbf{PS} \\
				\midrule
				\cellcolor{cat1color!12}\textbf{Moirai-Base} & \textbf{$-$13.36\%} & $-$12.93\% & $-$12.16\% & \textbf{$-$4.77\%} & $-$4.58\% & $-$4.08\% & \textbf{$-$5.79\%} & $-$5.66\% & $-$4.95\% \\
				\cellcolor{cat1color!12}\textbf{Moirai-Large} & \textbf{$-$12.44\%} & $-$11.78\% & $-$9.08\% & \textbf{$-$0.52\%} & $-$0.29\% & $+$0.95\% & \textbf{$-$1.57\%} & $-$1.46\% & $-$0.49\% \\
				\cellcolor{cat1color!12}\textbf{Moirai-MoE-Base} & \textbf{$-$19.72\%} & $-$19.62\% & $-$13.91\% & \textbf{$-$5.83\%} & $-$5.77\% & $-$4.09\% & \textbf{$-$5.51\%} & $-$5.43\% & $-$4.21\% \\
				\cellcolor{cat1color!12}\textbf{Moirai-MoE-Small} & $-$18.77\% & \textbf{$-$19.05\%} & $-$16.24\% & $-$7.78\% & \textbf{$-$7.82\%} & $-$7.07\% & \textbf{$-$5.62\%} & $-$5.58\% & $-$4.92\% \\
				\cellcolor{cat1color!12}\textbf{Moirai-Small} & $-$9.29\% & \textbf{$-$9.45\%} & $-$7.89\% & \textbf{$-$3.55\%} & $-$3.44\% & $-$2.30\% & $-$2.23\% & \textbf{$-$2.32\%} & $-$1.59\% \\
				\midrule
				\cellcolor{cat2color!12}\textbf{Chronos-T5-Base} & \textbf{$-$62.23\%} & $-$62.12\% & $-$60.25\% & \textbf{$-$39.04\%} & $-$38.96\% & $-$37.90\% & \textbf{$-$32.40\%} & $-$32.30\% & $-$31.56\% \\
				\cellcolor{cat2color!12}\textbf{Chronos-T5-Small} & \textbf{$-$60.38\%} & $-$60.29\% & $-$58.23\% & \textbf{$-$38.53\%} & $-$38.48\% & $-$37.45\% & \textbf{$-$31.14\%} & $-$31.06\% & $-$30.24\% \\
				\cellcolor{cat2color!12}\textbf{TabPFN-TS} & $-$21.20\% & \textbf{$-$21.21\%} & $-$17.18\% & \textbf{$-$10.67\%} & $-$10.57\% & $-$9.61\% & \textbf{$-$10.01\%} & $-$9.92\% & $-$9.25\% \\
				\midrule
				\cellcolor{cat3color!12}\textbf{Chronos-2} & $-$40.08\% & \textbf{$-$40.17\%} & $-$39.83\% & $-$17.30\% & $-$17.17\% & \textbf{$-$18.21\%} & \textbf{$-$14.79\%} & $-$14.66\% & $-$14.73\% \\
				\cellcolor{cat3color!12}\textbf{Chronos-Bolt-Base} & \textbf{$-$12.36\%} & $-$11.89\% & $-$8.42\% & \textbf{$-$3.87\%} & $-$3.48\% & $-$3.17\% & \textbf{$-$4.54\%} & $-$4.39\% & $-$4.16\% \\
				\cellcolor{cat3color!12}\textbf{Chronos-Bolt-Small} & \textbf{$-$24.45\%} & $-$24.03\% & $-$19.27\% & \textbf{$-$8.93\%} & $-$8.64\% & $-$7.83\% & \textbf{$-$6.79\%} & $-$6.71\% & $-$5.91\% \\
				\cellcolor{cat3color!12}\textbf{TiRex} & \textbf{$-$16.31\%} & $-$15.85\% & $-$14.55\% & $-$4.82\% & $-$4.64\% & \textbf{$-$5.16\%} & \textbf{$-$6.46\%} & $-$6.42\% & $-$6.12\% \\
				\cellcolor{cat3color!12}\textbf{TimesFM-2.5-200M} & \textbf{$-$51.33\%} & $-$51.14\% & $-$45.27\% & \textbf{$-$29.13\%} & $-$29.00\% & $-$26.38\% & \textbf{$-$26.33\%} & $-$26.21\% & $-$23.94\% \\
				\midrule
				\cellcolor{cat4color!12}\textbf{Sundial-Base} & $-$57.58\% & \textbf{$-$57.60\%} & $-$57.16\% & \textbf{$-$39.44\%} & \textbf{$-$39.44\%} & $-$39.22\% & $-$33.85\% & \textbf{$-$33.86\%} & $-$33.68\% \\
				\bottomrule
			\end{tabular}%
		}
		\caption{Relative change of the calibration errors after post-hoc recalibration under the chronological split, i.e., the change of each model's mean error over the 42 dataset configurations. Negative values are improvements, and bold numbers mark the best recalibration method for each model and metric. With a single exception, all errors decrease, but the gains are largest for \( \mathrm{PCE}_{\mathrm{pool}} \) and decrease for \( \mathrm{PCE}_{\mathrm{avg}} \) and \( \mathrm{QCE}_{\mathrm{avg}} \), highlighting again that recalibration via CDF relabeling mainly targets probabilistic miscalibration.}
		\label{tab:recalibration}
		\vspace{-0.2cm}
	\end{table*}
	
	\paragraph{Additional experiments.} The appendix complements these results with several further analyses. Simulations with independent series confirm that the significance tests of both gaps hold their nominal level and reliably separate horizon-dependent from value-dependent miscalibration (Appendix~\ref{app:significance}). Further case studies show that the pattern of Figure~\ref{fig:case_study} is no isolated example as across models and datasets, forecasts tend to be too extreme in both directions, with errors that cancel in average coverage. They also illustrate the complementary failure mode, where large coverage errors across the hours of the day cancel under pooling and keep recalibration above the floor of Theorem~\ref{thm:global_relabelling}(ii) (Appendix~\ref{app:case_studies}). When the calibration windows are drawn at random rather than from the past, recalibration is even more effective, with IR reducing \(\mathrm{PCE}_{\mathrm{pool}}\) by 67\% instead of 37\% (Appendix~\ref{app:additional_recal}). The recalibration methods perform similarly because they estimate the same empirical coverage curve at the fitted levels, and they differ mainly when extrapolating to tail quantiles, where Platt scaling performs best (Appendix~\ref{app:relabeling}). Finally, our conclusions are robust to tail quantiles, the number of Monte Carlo samples, scale differences between series, and point masses, while longer horizons enlarge the horizon gap (Appendix~\ref{app:robustness}).

	\section{Discussion and Conclusion}
	\label{sec:conclusion}
	
	We presented a calibration evaluation framework for probabilistic TSFMs that goes beyond pooled average coverage and enables a more fine-grained diagnosis of uncertainty reliability in multi-horizon forecasting. Our results show that standard pooled metrics can systematically understate miscalibration by masking both differences across forecast steps and additional value-dependent errors. Across 14 models and 55 datasets, we find that such miscalibration is widespread in current TSFMs, regardless of how predictive uncertainty is represented. We further studied post-hoc recalibration as a lightweight practical remedy. Both our analysis and experiments show that standard recalibration methods can substantially improve weaker average notions of calibration, also when they are fitted on past data only. They are less effective at removing stronger conditional deficiencies. The main limitations of our study lie in its scope as we focus only on horizon-wise marginal calibration and simple post-hoc recalibration methods. Moreover, our isotonic estimate of quantile-conditioned calibration is conservative, and point masses in the outcomes can inflate coverage errors. Extending the analysis to joint forecast calibration across time \cite{allen2024assessing}, more expressive recalibration methods that condition on the forecast horizon or on the forecast itself \cite{song2019distributioncalibration, kuleshov2022calibrated}, and training-time approaches for improving calibration \cite{dheur2024probabilistic} are promising directions for future work. Overall, we hope that the framework introduced here helps make calibration evaluation a more standard and informative part of assessing probabilistic TSFMs in practice.

	\bibliography{references}
	\bibliographystyle{plain}
	
	%
	\clearpage
	\appendix
	
	\section{Theoretical Results}
	\label{app:proofs}
	
	This appendix proves the calibration hierarchy (Proposition~\ref{thm:hierarchy}) and the limits of global relabeling (Theorem~\ref{thm:global_relabelling}), and characterizes the population target of the isotonic estimator (Lemma~\ref{lem:isotonic_target}). All arguments rest on two elementary observations. First, the absolute value of an average never exceeds the average of absolute values, and the difference measures how much positive and negative terms cancel. Each step of the hierarchy is an instance of this principle, applied to cancellation across horizons or across predicted values. Second, a single relabeling map moves the predicted quantile at a given level in the same direction for every horizon and every forecast. It can therefore remove errors that all horizons and forecasts share, but not errors that differ between them.
	
	\paragraph{Notation.}
	For a level \(t\in[0,1]\), let \(\hat q_{h,t}:=\hat F^{-1}_{Y_h\mid\mathbf X}(t)\) denote the predicted \(t\)-quantile at horizon \(h\), with the conventions \(\hat q_{h,0}=-\infty\) and \(\hat q_{h,1}=+\infty\). The two central quantities are the coverage at horizon \(h\) and the pooled coverage,
	\[
	p_h(t):=P\bigl(Y_h\le \hat q_{h,t}\bigr),\qquad \bar p(t):=\frac1H\sum_{h=1}^H p_h(t).
	\]
	Since the predicted quantiles are non-decreasing in \(t\), so are \(p_h\) and \(\bar p\), with value \(0\) at \(t=0\) and \(1\) at \(t=1\). At a grid level \(\tau_q\), we write
	\[
	r_{h,q}:=p_h(\tau_q)-\tau_q,
	\qquad
	C_{h,q}:=P\bigl(Y_h\le\hat q_{h,\tau_q}\mid\hat q_{h,\tau_q}\bigr)
	\]
	for the signed coverage error of horizon \(h\), which is positive for over-coverage and negative for under-coverage, and for the conditional coverage given the predicted value. In this notation, the calibration errors of Section~\ref{sec:calibration_framework} read
	\[
	\mathrm{PCE}_{\mathrm{pool}}=\frac1Q\sum_{q=1}^Q\Bigl|\frac1H\sum_{h=1}^H r_{h,q}\Bigr|,\qquad
	\mathrm{PCE}_{\mathrm{avg}}=\frac1{QH}\sum_{q=1}^Q\sum_{h=1}^H|r_{h,q}|,
	\]
	\[
	\mathrm{QCE}_{\mathrm{avg}}=\frac1{QH}\sum_{q=1}^Q\sum_{h=1}^H\mathbb E\,|C_{h,q}-\tau_q|.
	\]
	Recalibration with a map \(g\) replaces \(\hat q_{h,\tau}\) by \(\hat q_{h,g(\tau)}\) and keeps the nominal level \(\tau\). The recalibrated errors are therefore obtained by evaluating the coverages at \(g(\tau_q)\) instead of \(\tau_q\),
	\[
	\mathrm{PCE}^g_{\mathrm{pool}}=\frac1Q\sum_{q=1}^Q\bigl|\bar p(g(\tau_q))-\tau_q\bigr|,\qquad
	\mathrm{PCE}^g_{\mathrm{avg}}=\frac1{QH}\sum_{q=1}^Q\sum_{h=1}^H\bigl|p_h(g(\tau_q))-\tau_q\bigr|,
	\]
	and \(\mathrm{QCE}^g_{\mathrm{avg}}\) is defined analogously with the conditional coverage \(C^g_{h,q}:=P(Y_h\le\hat q_{h,g(\tau_q)}\mid\hat q_{h,g(\tau_q)})\).
	
	\subsection{Proof of the calibration error hierarchy}
	\label{app:proof_hierarchy}
	
	\paragraph{Intuition.}
	Each step of the chain removes one opportunity for errors to cancel. \(\mathrm{PCE}_{\mathrm{pool}}\) averages the signed errors over horizons before taking absolute values, so over-coverage at one horizon can offset under-coverage at another. \(\mathrm{PCE}_{\mathrm{avg}}\) takes absolute values per horizon and rules out this cancellation, but still averages over all forecasts of a horizon. \(\mathrm{QCE}_{\mathrm{avg}}\) additionally conditions on the predicted value and thereby also rules out cancellation between forecasts with different predicted values. \(\mathrm{QCE}_{\max}\) finally replaces the average over horizons by the worst horizon.
	
	\begin{proof}[Proof of Proposition~\ref{thm:hierarchy}]
		We verify the three inequalities in turn.
		
		\emph{Pooling over horizons.} By the triangle inequality, at every level
		\[
		\Bigl|\frac1H\sum_{h=1}^H r_{h,q}\Bigr|\le\frac1H\sum_{h=1}^H|r_{h,q}|,
		\]
		and averaging over \(q\) gives \(\mathrm{PCE}_{\mathrm{pool}}\le\mathrm{PCE}_{\mathrm{avg}}\).
		
		\emph{Conditioning on the predicted value.} By the tower property, the conditional coverage averages to the marginal coverage, \(\mathbb E[C_{h,q}]=p_h(\tau_q)\). Jensen's inequality for the convex function \(x\mapsto|x-\tau_q|\) therefore gives
		\[
		|r_{h,q}|=\bigl|\mathbb E[C_{h,q}]-\tau_q\bigr|\le\mathbb E\,|C_{h,q}-\tau_q|,
		\]
		and averaging over \(q\) and \(h\) gives \(\mathrm{PCE}_{\mathrm{avg}}\le\mathrm{QCE}_{\mathrm{avg}}\).
		
		A mean over horizons never exceeds their maximum, so \(\mathrm{QCE}_{\mathrm{avg}}\le\mathrm{QCE}_{\max}\).
	\end{proof}
	
	\paragraph{The same chain holds for the estimators.}
	\phantomsection\label{rem:empirical_hierarchy}
	The estimators of Appendix~\ref{sec:calib_computation} satisfy the hierarchy exactly on every sample, so both estimated gaps are non-negative. The first and the last step carry over verbatim with the empirical coverages \(\hat p_h\) in place of \(p_h\). For the middle step, the isotonic fit assigns each block of consecutive forecasts the mean of their coverage indicators, so the fitted values average to the empirical coverage \(\hat p_h(\tau_q)\). This is the empirical counterpart of the tower property, and Jensen's inequality applies to the empirical distribution.
	
	\subsection{Proof of the limits of quantile relabeling}
	\label{app:proof_recalibration}
	
	We prove the three parts of Theorem~\ref{thm:global_relabelling} separately and precede each proof by its main idea.
	
	\paragraph{Part (i): pooled calibration can always be restored.}
	To make the pooled coverage at nominal level \(\tau\) equal to \(\tau\), it suffices to find a model level \(t\) at which the pooled coverage equals \(\tau\) and to relabel \(\tau\) to \(t\). Since \(\bar p\) goes from \(0\) to \(1\), such a level exists whenever \(\bar p\) has no jumps, and the generalized inverse selects the smallest one. Isotonic recalibration implements this idea by estimating \(\bar p\) on calibration data.
	
	\begin{proof}[Proof of Theorem~\ref{thm:global_relabelling}(i)]
		We start by defining \(g^\star\) as the generalized inverse of \(\bar p\).  For \(\tau\in(0,1)\) let
		\[
		g^\star(\tau):=\bar p^{-1}(\tau)=\inf\{t\in[0,1]:\bar p(t)\ge\tau\},
		\]
		and set \(g^\star(0)=0\) and \(g^\star(1)=1\). Fix \(\tau\in(0,1)\) and write \(t^\star:=g^\star(\tau)\). The set \(\{t:\bar p(t)\ge\tau\}\) contains \(1\) and is closed because \(\bar p\) is continuous, so \(\bar p(t^\star)\ge\tau\). Since \(\bar p(0)=0<\tau\), we have \(t^\star>0\) and \(\bar p(t)<\tau\) for all \(t<t^\star\), and letting \(t\uparrow t^\star\) gives \(\bar p(t^\star)\le\tau\) by continuity. Hence
		\[
		\bar p\bigl(g^\star(\tau_q)\bigr)=\tau_q\qquad\text{for every grid level } \tau_q,
		\]
		and \(\mathrm{PCE}^{g^\star}_{\mathrm{pool}}=0\). Moreover, \(g^\star\) is non-decreasing because the set shrinks as \(\tau\) grows, so the recalibrated quantiles \(\hat q_{h,g^\star(\tau)}\) remain ordered in \(\tau\).
	\end{proof}
	Continuity of \(\bar p\) is needed only for this part, to solve \(\bar p(t)=\tau_q\) exactly. If \(\bar p\) jumps over a target level, for example for predictive distributions with point masses or tied quantiles, no deterministic relabeling attains exact pooled coverage at that level. For quantile-only models, the statement refers to the forecast whose quantile function is interpolated between the grid levels. Alternatively, one can keep the predicted values and only relabel their nominal levels as \(\bar p(\tau_q)\), the form used in the experiments (see below). The pooled error is then zero without any continuity assumption, but the levels are no longer the requested ones. Parts (ii) and (iii) hold without this assumption.
	
	\paragraph{Part (ii): cancellation across horizons cannot be undone.}
	At a given level, a global map selects one model level \(t\) for all horizons. Raising \(t\) raises the coverage of every horizon, and lowering it lowers all of them. If some horizons over-cover while others under-cover, every such shift brings one group closer to the nominal level only at the expense of the other, or leaves it unchanged. The horizon gap measures exactly how much over- and under-coverage coexist at the same level, which yields the floor.
	
	\begin{proof}[Proof of Theorem~\ref{thm:global_relabelling}(ii)]
		For each level, collect the total over- and under-coverage across horizons,
		\[
		S^+_q:=\sum_{h=1}^H\max(r_{h,q},0),\qquad S^-_q:=\sum_{h=1}^H\max(-r_{h,q},0).
		\]
		
		\emph{Step 1: the horizon gap measures cancellation.} Since \(\sum_h|r_{h,q}|=S^+_q+S^-_q\), \(|\sum_h r_{h,q}|=|S^+_q-S^-_q|\), and \(a+b-|a-b|=2\min(a,b)\),
		\[
		\Delta_H=\mathrm{PCE}_{\mathrm{avg}}-\mathrm{PCE}_{\mathrm{pool}}=\frac1Q\sum_{q=1}^Q\frac1H\Bigl(S^+_q+S^-_q-|S^+_q-S^-_q|\Bigr)=\frac2{QH}\sum_{q=1}^Q\min(S^+_q,S^-_q).
		\tag{$\ast$}
		\]
		Thus \(\Delta_H\) is twice the amount of over-coverage that is offset by under-coverage at other horizons, averaged over levels.
		
		\emph{Step 2: a global map cannot correct both sides.} Fix any map \(g\) and a level \(q\), and let \(t:=g(\tau_q)\). Since every \(p_h\) is non-decreasing,
		\[
		t\ge\tau_q\;\Longrightarrow\;p_h(t)-\tau_q\ge r_{h,q}\ \text{ for all } h,
		\qquad
		t\le\tau_q\;\Longrightarrow\;p_h(t)-\tau_q\le r_{h,q}\ \text{ for all } h.
		\]
		In the first case, every over-covering horizon still over-covers by at least \(r_{h,q}\), so \(\sum_h|p_h(t)-\tau_q|\ge S^+_q\). In the second case, every under-covering horizon still under-covers by at least \(|r_{h,q}|\), so \(\sum_h|p_h(t)-\tau_q|\ge S^-_q\). In both cases,
		\[
		\sum_{h=1}^H\bigl|p_h(g(\tau_q))-\tau_q\bigr|\ge\min(S^+_q,S^-_q).
		\]
		
		\emph{Step 3: averaging.} Dividing by \(QH\), summing over \(q\), and using \((\ast)\) gives \(\mathrm{PCE}^g_{\mathrm{avg}}\ge\Delta_H/2\).
		
		\emph{Sharpness.} Let \(H=2\), \(Q=1\), \(\tau_1=\tfrac12\), and \(a\in(0,\tfrac14)\). Take uniform predictive distributions, \(\hat q_{h,t}=t\), and outcomes \(Y_h\) on \([0,1]\) with continuous CDFs \(c_h\), so that \(p_h=c_h\). Let \(c_1\) interpolate \((0,0),(\tfrac12,\tfrac12+a),(\tfrac12+a,\tfrac12+a),(1,1)\) and \(c_2\) interpolate \((0,0),(\tfrac12,\tfrac12-a),(\tfrac12+a,\tfrac12),(1,1)\) piecewise linearly. Then \(r_{1,1}=a\), \(r_{2,1}=-a\), and \(\Delta_H=a\) by \((\ast)\). Relabeling to \(g(\tfrac12)=\tfrac12+a\) calibrates the second horizon exactly, while the coverage of the first horizon is flat on \([\tfrac12,\tfrac12+a]\) and its error stays at \(a\). Hence \(\mathrm{PCE}^g_{\mathrm{avg}}=\tfrac12(a+0)=\Delta_H/2\), and the bound is attained.
	\end{proof}
	
	\paragraph{Forward relabeling, as used in the experiments.}
	\phantomsection\label{rem:forward}
	In the experiments, recalibrated forecasts are represented in an alternative form: the predicted value \(\hat q_{h,\tau_q}\) is kept and receives a new nominal level \(\ell_q\) shared by all horizons, with error \(\frac1{QH}\sum_{q,h}|p_h(\tau_q)-\ell_q|\). This avoids inverting a sparse quantile grid, but its averages over a finite grid need not coincide with those of the transformed quantiles \(\hat q_{h,g(\tau_q)}\). The bound of part (ii) holds for this form as well, by the same argument with the roles of the two cases exchanged. If \(\ell_q\ge\tau_q\), every under-covering horizon contributes at least \(|r_{h,q}|\), and otherwise every over-covering horizon contributes at least \(r_{h,q}\). Because the argument applies to any distribution, including the empirical distribution of a test set, the bound also holds exactly on every sample.
	
	\paragraph{Part (iii): restoring pooled coverage can expose conditional errors.}
	Quantile-conditioned calibration conditions on the predicted quantile. If all forecasts predict the same quantile, conditioning on it has no effect and \(\mathrm{QCE}\) reduces to the marginal error. The construction exploits this. Two regimes predict the same median, so that their very different calibration is invisible before recalibration. Every map that restores pooled coverage must move to a level at which the two regimes predict different quantiles. The relabeled quantile then reveals the regime, and \(\mathrm{QCE}\) exposes that the model over-covers in one regime and under-covers in the other. Recalibration thus removes the pooled error while making the conditional error almost as large as possible.
	
	\begin{proof}[Proof of Theorem~\ref{thm:global_relabelling}(iii)]
		Fix \(\varepsilon\in(0,\tfrac12)\) and \(\eta\in(0,\tfrac12-\varepsilon)\). Let \(H=Q=1\), \(\tau_1=\tfrac12\), and let a latent regime \(X\) be uniform on \(\{A,B\}\). Table~\ref{tab:construction} specifies the predictive distributions, their quantile functions \(q_x\), the outcome distributions, and the resulting regime-wise coverages \(c_x(t):=P(Y\le q_x(t)\mid X=x)\). Both coverages are continuous, so \(\bar p=\tfrac12(c_A+c_B)\) is continuous.
		
		\begin{table}[htb]
			\centering
			\small
			\renewcommand{\arraystretch}{1.6}
			\resizebox{\textwidth}{!}{%
				\begin{tabular}{@{} c c c c c @{}} \toprule
					\textbf{Regime} & \textbf{Predictive distribution} & \textbf{Quantile function} \(q_x(t)\) & \textbf{Outcome} \(Y\mid X=x\) & \textbf{Coverage} \(c_x(t)\) \\
					\midrule
					\(A\) & \(\tfrac12\mathrm{U}[-2,0]+\tfrac12\mathrm{U}[0,1]\) & \(\begin{cases}4t-2,&t\le\tfrac12\\2t-1,&t>\tfrac12\end{cases}\) & \(\mathrm{U}[-2,-1]\) & \(\min(4t,1)\) \\
					\(B\) & \(\tfrac12\mathrm{U}[-1,0]+\tfrac12\mathrm{U}[1,2]\) & \(\begin{cases}2t-1,&t\le\tfrac12\\2t,&t>\tfrac12\end{cases}\) & \(\begin{array}{c}2\eta\,\mathrm{U}[-\tfrac14,0]\\+\,(1-2\eta)\,\mathrm{U}[1,2]\end{array}\) & \(\begin{cases}0,&t\le\tfrac38\\2\eta(8t-3),&\tfrac38<t\le\tfrac12\\2\eta+(1-2\eta)(2t-1),&t>\tfrac12\end{cases}\) \\
					\bottomrule
			\end{tabular}}
			\caption{Construction for Theorem~\ref{thm:global_relabelling}(iii). Both regimes predict the median \(q_A(\tfrac12)=q_B(\tfrac12)=0\), but the model over-covers in regime \(A\) and under-covers in regime \(B\).}
			\label{tab:construction}
		\end{table}
		
		\emph{Step 1: before recalibration, the regimes are indistinguishable.} Both regimes predict the median \(q_A(\tfrac12)=q_B(\tfrac12)=0\). The predicted median is therefore constant, its conditional coverage equals the marginal coverage \(\tfrac12\bigl(c_A(\tfrac12)+c_B(\tfrac12)\bigr)=\tfrac12(1+2\eta)\), and
		\[
		\mathrm{PCE}_{\mathrm{pool}}=\mathrm{QCE}_{\mathrm{avg}}=\eta.
		\]
		
		\emph{Step 2: restoring pooled coverage forces a level that separates the regimes.} The pooled coverage satisfies \(\bar p(t)=2t<\tfrac12\) on \([0,\tfrac14)\), \(\bar p=\tfrac12\) on \([\tfrac14,\tfrac38]\), and \(\bar p>\tfrac12\) beyond. Hence
		\[
		\mathrm{PCE}^g_{\mathrm{pool}}=\bigl|\bar p(g(\tfrac12))-\tfrac12\bigr|=0
		\quad\Longleftrightarrow\quad
		t:=g(\tfrac12)\in[\tfrac14,\tfrac38],
		\]
		and the map of part (i) gives \(t=\tfrac14\). For every such \(t\), the relabeled medians \(q_A(t)=4t-2\) and \(q_B(t)=2t-1\) differ because \(t\neq\tfrac12\).
		
		\emph{Step 3: the relabeled median reveals the regime.} Conditional on the relabeled median, the regime is known, and the conditional coverages are \(c_A(t)=1\) under \(A\) and \(c_B(t)=0\) under \(B\). Thus
		\[
		\mathrm{QCE}^{g}_{\mathrm{avg}}=\tfrac12\bigl|1-\tfrac12\bigr|+\tfrac12\bigl|0-\tfrac12\bigr|=\tfrac12
		\]
		for every \(g\) with \(\mathrm{PCE}^g_{\mathrm{pool}}=0\), and the increase is \(\tfrac12-\eta>\varepsilon\).
	\end{proof}
	
	\subsection{Population target of the isotonic estimator}
	
	The isotonic estimator of Appendix~\ref{sec:calib_computation} does not target the conditional coverage curve itself, but its best non-decreasing approximation. This approximation pools adjacent predicted values into blocks and averages the conditional coverage within each block. Averaging can only pull values towards their mean, so the resulting error lies between the marginal and the full conditional error. If the conditional coverage is already non-decreasing, no pooling is needed and the target equals \(\mathrm{QCE}_h\).
	
	\begin{lemma}
		\label{lem:isotonic_target}
		Fix \(h\) and \(q\), write \(Z:=\hat q_{h,\tau_q}\) and \(c(z):=P(Y_h\le Z\mid Z=z)\), and let \(c^{\uparrow}\) be the \(L^2(P_Z)\)-projection of \(c\) onto non-decreasing functions, i.e., the population counterpart of the isotonic fit in Appendix~\ref{sec:calib_computation}. Then
		\[
		|p_h(\tau_q)-\tau_q|\;\le\;\mathbb E\,|c^{\uparrow}(Z)-\tau_q|\;\le\;\mathbb E\,|c(Z)-\tau_q|=\mathrm{QCE}_h(\tau_q),
		\]
		with equality on the right if \(c\) is non-decreasing.
	\end{lemma}
	
	\begin{proof}
		The isotonic projection is the conditional expectation of \(c\) given its own level sets \cite{barlow1972isotonic,robertson1988order},
		\[
		\mathbb E\bigl[c(Z)\mid c^{\uparrow}(Z)\bigr]=c^{\uparrow}(Z)\quad\text{almost surely}.
		\]
		Taking expectations gives \(\mathbb E[c^{\uparrow}(Z)]=\mathbb E[c(Z)]=p_h(\tau_q)\), and the left inequality is Jensen's inequality. The right inequality is the conditional Jensen inequality
		\[
		|c^{\uparrow}(Z)-\tau_q|=\bigl|\mathbb E[c(Z)-\tau_q\mid c^{\uparrow}(Z)]\bigr|\le\mathbb E\bigl[|c(Z)-\tau_q|\;\big|\;c^{\uparrow}(Z)\bigr],
		\]
		followed by taking expectations. If \(c\) is non-decreasing, then \(c^{\uparrow}=c\) and equality holds.
	\end{proof}
	
	The lower bound can be strict. Let \(Z\) take the values \(-1\) and \(0\) with probability \(\tfrac12\) each, with conditional coverages \(c(-1)=1\) and \(c(0)=0\) at level \(\tau_q=\tfrac12\). Then \(\mathrm{QCE}_h(\tau_q)=\tfrac12\), whereas the projection pools both values into the constant \(\tfrac12\), so the isotonic target is zero, like the marginal error. The diagnostic therefore detects value dependence in which coverage increases with the predicted value. This is the pattern of forecasts that are too extreme in both directions, which dominates our case studies (Appendix~\ref{app:case_studies}), while decreasing or non-monotone patterns may be missed.
	
	\section{Experimental Details}
	\label{app:experimental_details}
	
	\subsection{Models and compute}
	All model inference was performed on a single NVIDIA RTX A5000 GPU running Linux. Models were loaded via HuggingFace \texttt{transformers} or the model-specific package and evaluated sequentially per dataset with batch size 32. We evaluate 14 foundation models spanning the four uncertainty quantification categories; Table~\ref{tab:models} lists the model identifiers. Code implementing all metrics, significance tests, and recalibration methods is released at \texttt{https://github.com/thomdeck/tsfm-calibration}.
	
	\begin{table}[H]
		\centering
		\caption{Models evaluated, grouped by uncertainty quantification category.}
		\label{tab:models}
		\small
		\begin{tabular}{lll}
			\toprule
			\textbf{UQ Category} & \textbf{Model} & \textbf{HuggingFace ID / Package}  \\
			\midrule
			\rowcolor{cat1color!12}Cat.\ 1: Parametric       & Moirai-Small      & \texttt{Salesforce/moirai-1.1-R-small}  \\
			\rowcolor{cat1color!12}& Moirai-Base       & \texttt{Salesforce/moirai-1.1-R-base}   \\
			\rowcolor{cat1color!12}& Moirai-Large      & \texttt{Salesforce/moirai-1.1-R-large}  \\
			\rowcolor{cat1color!12}& Moirai-MoE-Small  & \texttt{Salesforce/moirai-moe-1.0-R-small}  \\
			\rowcolor{cat1color!12}& Moirai-MoE-Base   & \texttt{Salesforce/moirai-moe-1.0-R-base}   \\
			\midrule
			\rowcolor{cat2color!12}Cat.\ 2: Discretized      & Chronos-T5-Small  & \texttt{amazon/chronos-t5-small}   \\
			\rowcolor{cat2color!12}& Chronos-T5-Base   & \texttt{amazon/chronos-t5-base}    \\
			\rowcolor{cat2color!12}& TabPFN-TS         & PriorLabs/tabpfn-time-series     \\
			\midrule
			\rowcolor{cat3color!12}Cat.\ 3: Direct Quantile  & Chronos-2          & \texttt{amazon/chronos-2}          \\
			\rowcolor{cat3color!12}& Chronos-Bolt-Small & \texttt{amazon/chronos-bolt-small}  \\
			\rowcolor{cat3color!12}& Chronos-Bolt-Base  & \texttt{amazon/chronos-bolt-base}   \\
			\rowcolor{cat3color!12}& TiRex             & \texttt{NX-AI/TiRex}               \\
			\rowcolor{cat3color!12}& TimesFM-2.5-200M  & \texttt{google/timesfm-2.5-200m-pytorch} \\
			\midrule
			\rowcolor{cat4color!12}Cat.\ 4: Generative       & Sundial-Base      & \texttt{thuml/sundial-base-128m}   \\
			\bottomrule
		\end{tabular}
	\end{table}
	
	\noindent\textbf{Quantile levels.}
	All probabilistic forecasts are evaluated at the nine levels $\tau \in \{0.1, 0.2, \ldots, 0.9\}$, the intersection of the levels natively supported by all models.
	
	\noindent\textbf{Sampling-based models.}
	For Chronos-T5 (autoregressive categorical sampling over 4096 bins), Moirai~1.1 and Moirai-MoE (a mixture of Student-$t$, Normal, Negative Binomial, and Log-Normal distributions), and Sundial (flow matching, 128M parameters), we draw $S=100$ sample paths per forecast and read off empirical quantiles.
	
	\noindent\textbf{Quantile-output models.}
	TabPFN-TS applies the TabPFN-v2 tabular foundation model \cite{hollmann2025accurate} to time series by turning forecasting into a tabular regression task. It represents its predictive distribution as a bar distribution over a fixed grid and returns the nine quantiles in a single forward pass. Chronos-2, Chronos-Bolt, TiRex, and TimesFM output the nine quantile levels directly.
	
	\subsection{Data}
	\label{app:data}
	We use the GIFT-Eval benchmark \cite{aksu2024gift} (\texttt{Salesforce/gift-eval} on HuggingFace), which covers energy, traffic, weather, finance, health, sales, and cloud infrastructure data. The calibration analysis uses the \emph{short-term} task of all 55 dataset configurations (770 model--dataset pairs). The recalibration experiments require a calibration and a test split, and we restrict them to the 42 configurations with at least 50 forecast windows. Medium- and long-term tasks are used only for the horizon-length analysis in Appendix~\ref{app:robustness}. All models are evaluated on the same forecast windows of each dataset, so that model comparisons are paired and based on equal sample sizes. Missing target values (present in ten configurations) carry no calibration information and are excluded from all computations; they are never counted as non-covered.
	
	\subsection{Estimating the calibration errors}
	\label{sec:calib_computation}
	
	For forecast window \(i\), horizon \(h\), and level \(\tau\), let \(z_{i,h}(\tau)=\mathbf 1\{y_{i,h}\le\hat q_{i,h,\tau}\}\) be the coverage indicator, and let \(\hat p_h(\tau)\) be its mean over the \(N_h\) windows with an observed target at horizon \(h\).
	
	\paragraph{Probabilistic calibration errors.}
	The estimators replace \(p_h\) by \(\hat p_h\) in the definitions of Section~\ref{sec:calibration_framework}:
	\(\widehat{\mathrm{PCE}}_{\mathrm{pool}}=\frac1Q\sum_q|\frac1H\sum_h\hat p_h(\tau_q)-\tau_q|\),
	\(\widehat{\mathrm{PCE}}_h=\frac1Q\sum_q|\hat p_h(\tau_q)-\tau_q|\),
	\(\widehat{\mathrm{PCE}}_{\mathrm{avg}}=\frac1H\sum_h\widehat{\mathrm{PCE}}_h\), and
	\(\widehat{\mathrm{PCE}}_{\max}=\max_h\widehat{\mathrm{PCE}}_h\), i.e., the worst horizon after averaging over levels.
	
	\paragraph{Quantile-conditioned calibration errors.}
	For each horizon \(h\) and level \(\tau_q\), we estimate the conditional coverage curve \(c_{h,q}(x)=P(Y_h\le\hat q_{h,\tau_q}\mid\hat q_{h,\tau_q}=x)\) by least-squares isotonic (non-decreasing) regression of the indicators \(z_{i,h}(\tau_q)\) on the predicted values \(\hat q_{i,h,\tau_q}\), computed with the pool-adjacent-violators algorithm. Windows with tied predicted values share one fitted value. With the fitted curve \(\hat c_{h,q}\),
	\[
	\widehat{\mathrm{QCE}}_h(\tau_q)=\frac1{N_h}\sum_{i}\bigl|\hat c_{h,q}(\hat q_{i,h,\tau_q})-\tau_q\bigr|,
	\]
	which is averaged over levels and then over horizons (\(\widehat{\mathrm{QCE}}_{\mathrm{avg}}\)) or maximized over horizons (\(\widehat{\mathrm{QCE}}_{\max}\)). The gaps are \(\widehat\Delta_H=\widehat{\mathrm{PCE}}_{\mathrm{avg}}-\widehat{\mathrm{PCE}}_{\mathrm{pool}}\) and \(\widehat\Delta_C=\widehat{\mathrm{QCE}}_{\mathrm{avg}}-\widehat{\mathrm{PCE}}_{\mathrm{avg}}\). Each curve is fitted within one dataset and model.
	
	\paragraph{Why this estimator.}
	\begin{itemize}[leftmargin=12pt, itemsep=1pt]
		\item \emph{No tuning.} Isotonic regression chooses the grouping of forecasts automatically and requires no bins or bandwidth. This is the principle behind CORP reliability diagrams \cite{dimitriadis2021stable,gneiting2023regression}, which use it for the same reason. The same principle underlies the T-PAV decomposition of the quantile score \cite{gneiting2023regression}, which measures miscalibration in units of the outcome through the pinball loss; QCE measures it in coverage probability and therefore remains comparable across datasets and series of different scale.
		\item \emph{Same units as PCE, exact hierarchy.} \(\widehat{\mathrm{QCE}}\) is measured in coverage probability, and the empirical hierarchy \(\widehat{\mathrm{PCE}}_{\mathrm{pool}}\le\widehat{\mathrm{PCE}}_{\mathrm{avg}}\le\widehat{\mathrm{QCE}}_{\mathrm{avg}}\le\widehat{\mathrm{QCE}}_{\max}\) holds on every sample (Appendix~\ref{rem:empirical_hierarchy}), so both estimated gaps are non-negative.
		\item \emph{Comparable across datasets.} The estimate depends on the data only through the coverage indicators and the ordering of the predicted values. It is therefore unchanged by any strictly increasing transformation applied jointly to outcomes and forecasts (change of units, rescaling, log transform), and every forecast contributes equally regardless of the magnitude of its series. This matters for a benchmark of zero-shot models applied to arbitrary domains.
		\item \emph{Conservative population target.} The estimator targets the isotonic projection of \(c_{h,q}\). If \(c_{h,q}\) is non-decreasing, this is \(\mathrm{QCE}_h(\tau_q)\) itself; otherwise it lies between \(\mathrm{PCE}_h(\tau_q)\) and \(\mathrm{QCE}_h(\tau_q)\) (Lemma~\ref{lem:isotonic_target}), so the monotonicity constraint never overstates value dependence. Conversely, it can understate decreasing or non-monotone value dependence, which the estimate may therefore miss.
		\item \emph{Finite-sample bias.} Fitting and evaluating the curve on the same windows makes \(\widehat{\mathrm{QCE}}\) positive even without value dependence. We quantify this component with the permutation test of Appendix~\ref{app:significance}. Neither a small gap nor a non-significant test certifies conditional calibration.
	\end{itemize}
	
	\paragraph{Outcomes with point masses.}
	For continuous predictive CDFs, Definition~\ref{def:conditional_quantile_calibration} coincides with CEP calibration \cite{mason2007conditional,gneiting2023regression}. Both this definition and probabilistic calibration presuppose that the outcome has no point masses, i.e., no single value that occurs with positive probability. Some datasets violate this, for example through zero demand in intermittent sales or zero solar power at night. If a predicted quantile equals such a value \(a\), every outcome \(Y=a\) counts as covered, so coverage exceeds the nominal level even when the forecast equals the true conditional distribution. At night, for instance, the true distribution of solar power is a point mass at zero. All its quantiles are zero and are covered with probability one, so a perfect forecast has coverage error \(1-\tau\) at every level \(\tau\). Average coverage dilutes this effect with the remaining forecasts, whereas quantile conditioning isolates exactly the forecasts whose quantile sits at the point mass, so such datasets produce the largest conditioning gaps. We therefore flag them and check that they do not drive our conclusions (Appendix~\ref{app:robustness}).
	
	\subsection{Testing the diagnostic gaps}
	\label{app:significance}
	
	\paragraph{Why a test is needed.}
	Both gaps are estimated and evaluated on the same data, and their estimates are positive even when coverage depends neither on the predicted value nor on the horizon as the isotonic fit always finds some spurious structure, and per-horizon coverages always fluctuate. To decide whether an observed gap reflects real structure, we compare it with the distribution of the same statistic on data from which this structure has been removed while everything else, including the dependence between windows, horizons, and levels, is kept.
	
	\paragraph{Conditioning gap.}
	The null hypothesis is that the coverage indicators of a series are unrelated to its predicted values. We generate null data by randomly reassigning the complete coverage-indicator trajectories (all windows, horizons, and levels of a series) among series with the same number of windows, while keeping all predicted quantiles fixed. This leaves every per-horizon coverage, and hence every PCE, unchanged, preserves the temporal and cross-horizon dependence of the indicators, and removes only the association between coverage and predicted value. Recomputing \(\widehat\Delta_C\) on \(B=999\) such data sets gives its null distribution. The test is exact when series are exchangeable under the null hypothesis.
	
	\paragraph{Horizon gap.}
	The null hypothesis is that coverage is the same at all horizons. We resample whole series with replacement (a series-cluster bootstrap, which preserves arbitrary dependence within a series) and shift each bootstrap coverage curve to \(\hat p^*_h(\tau)-\hat p_h(\tau)+\frac1H\sum_{h'}\hat p_{h'}(\tau)\), so that all horizons share the observed pooled coverage. Recomputing \(\widehat\Delta_H\) on \(B=999\) bootstrap samples gives an approximate null distribution, which assumes that different series are independent.
	
	\paragraph{What we report.}
	For both gaps, the p-value is \((1+\#\{\text{null values}\ge\text{observed}\})/(B+1)\) \cite{phipson2010permutation}, and the \emph{corrected gap} is the observed gap minus the mean of its null distribution, i.e., the part of the gap not explained by estimation noise. We report the share of model--dataset pairs with \(p\le0.05\); if no pair had any structure, about 5\% would be significant by chance. A rejection indicates that coverage varies with the predicted value or the horizon in a way that produces cancellation, which is the structure the gaps are designed to reveal. Both tests require at least two series. For datasets consisting of a single series (\texttt{saugeenday} and \texttt{us\_births}, 84 model--dataset pairs), every bootstrap sample reproduces the original data and no nontrivial permutation exists, so we report only their descriptive metrics and exclude them from all significance shares. Over the remaining 686 pairs, 82\% of the horizon gaps and 65\% of the conditioning gaps are significant.
	
	\paragraph{Size and power.}
	We simulate 40 independent series with 10 windows each and \(H=24\), with Gaussian forecast errors that are AR(1) across horizons with correlation \(\rho\) and window-specific location and scale (300 replicates, \(B=199\)). In the calibrated scenario the predictive distribution equals the true one. In the horizon scenario the predictive spread is too wide at short and too narrow at long horizons, and in the value scenario it is too narrow for high and too wide for low predicted values. Table~\ref{tab:size} shows that both tests hold their nominal level up to simulation error (the \(\Delta_H\) test is slightly liberal under very strong cross-horizon correlation), detect the failure mode they are designed for, and reject the other one only at about the nominal rate, so the two tests separate the two failure modes. The \(\Delta_C\) test loses power under very strong cross-horizon correlation (\(\rho=0.9\)), where the 24 horizons carry little independent information.
	
	\begin{table}[htb]
		\centering
		\small
		\setlength{\tabcolsep}{5pt}
		\begin{tabular}{@{} l ccc ccc @{}} \toprule
			& \multicolumn{3}{c}{\(\Delta_C\) permutation test} & \multicolumn{3}{c}{\(\Delta_H\) bootstrap test} \\
			\cmidrule(lr){2-4} \cmidrule(l){5-7}
			\textbf{Scenario} & \(\rho=0\) & \(0.5\) & \(0.9\) & \(\rho=0\) & \(0.5\) & \(0.9\) \\
			\midrule
			Calibrated (size) & 0.05 & 0.04 & 0.04 & 0.06 & 0.04 & 0.08 \\
			Value-dependent spread & 1.00 & 0.99 & 0.48 & 0.03 & 0.06 & 0.06 \\
			Horizon-dependent spread & 0.03 & 0.06 & 0.05 & 1.00 & 1.00 & 1.00 \\
			\bottomrule
		\end{tabular}
		\caption{Rejection rates at level 0.05 in the simulation study. The calibrated scenario gives the size of each test; the other rows give the power against the failure mode the test targets and the rejection rate under the other one.}
		\label{tab:size}
	\end{table}
	
	\paragraph{Gaps by dataset size.}
	Table~\ref{tab:strata} stratifies the model--dataset pairs with at least two series by the number of forecast windows. The estimation-noise component of \(\widehat\Delta_C\) is large for small datasets (0.043 below 50 windows) and much smaller for larger ones, whereas the corrected gap does not shrink with dataset size and is largest for the largest datasets (0.024), as expected for real structure. Raw gaps of small datasets should therefore not be interpreted without the test, and the tests have little power there.
	
	\begin{table}[htb]
		\centering
		\small
		\begin{tabular}{@{} l r rrr rr @{}} \toprule
			& & \multicolumn{3}{c}{\(\widehat\Delta_C\)} & \multicolumn{2}{c}{\textbf{\% significant}} \\
			\cmidrule(lr){3-5} \cmidrule(l){6-7}
			\textbf{Windows} & \textbf{Pairs} & \textbf{raw} & \textbf{noise} & \textbf{corrected} & \(\Delta_C\) & \(\Delta_H\) \\
			\midrule
			$<50$ & 98 & 0.063 & 0.043 & 0.020 & 37\% & 61\% \\
			50--199 & 98 & 0.023 & 0.016 & 0.007 & 47\% & 68\% \\
			200--499 & 196 & 0.024 & 0.010 & 0.014 & 76\% & 94\% \\
			500--999 & 182 & 0.023 & 0.007 & 0.015 & 68\% & 80\% \\
			$\ge1000$ & 112 & 0.035 & 0.011 & 0.024 & 79\% & 95\% \\
			\bottomrule
		\end{tabular}
		\caption{Conditioning gap and share of significant gaps (\(p\le0.05\)) by number of forecast windows, over the 686 model--dataset pairs with at least two series. ``Noise'' is the mean of the permutation null distribution.}
		\label{tab:strata}
	\end{table}
	
	\subsection{Recalibration protocol}
	\label{sec:recalibration}
	
	\paragraph{Calibration and test splits.}
	Each dataset is split into 20\% calibration and 80\% test windows, and no window is used in both.
	\begin{itemize}[leftmargin=12pt, itemsep=1pt]
		\item \textbf{Chronological (main text).} For series with \(W\ge2\) windows, the first \(\lceil0.2W\rceil\) windows in time are used for calibration and the remaining ones for testing, so that calibration always precedes evaluation. Series with a single window cannot be split in time; for these, a random 20\% of the series is used for calibration (a cold-start deployment). Eleven of the 42 configurations consist only of single-window series (\texttt{m4\_*}, \texttt{hospital}, \texttt{covid\_deaths}, \texttt{car\_parts}, \texttt{restaurant}, \texttt{solar/W}).
		\item \textbf{Random (Appendix~\ref{app:additional_recal}).} A random permutation of the windows, which ignores temporal order and is reported for comparison.
	\end{itemize}
	
	\paragraph{Methods.}
	All four methods fit a single non-decreasing map from nominal to recalibrated levels after pooling the calibration windows and horizons. The predicted quantile values are left unchanged and receive the recalibrated levels, and all metrics after recalibration compare test coverage with these levels (Appendix~\ref{rem:forward}).
	\begin{itemize}[leftmargin=12pt, itemsep=1pt]
		\item \textbf{Isotonic Regression (IR)} \cite{kuleshov2018calibrated}: a non-decreasing map from each nominal level to its empirical coverage on the calibration windows, interpolated on a 1000-point grid.
		\item \textbf{Platt Scaling (PS)} \cite{platt1999probabilistic}: the map \(\sigma(a\,\mathrm{logit}(\tau)+b)\), fitted by least squares to the empirical calibration coverages.
		\item \textbf{Histogram Binning (HB)} \cite{zadrozny2002transforming}: the empirical CDF of the PIT values evaluated at the edges of \(B=20\) equal-width bins and interpolated linearly in between.
		\item \textbf{KDE Calibration (KDE)} \cite{dheur2023largescale}: the smoothed PIT CDF \(\hat F(\alpha)=\frac1N\sum_i\sigma(\tau_{\mathrm{bw}}(\alpha-u_i))\) with \(\tau_{\mathrm{bw}}=100\).
	\end{itemize}
	For HB and KDE, approximate PIT values are obtained by linear interpolation between the nine predicted quantiles of each forecast; values at or below the lowest quantile are set to \(10^{-6}\) and values above the highest to \(1-10^{-6}\). A target equal to a predicted quantile counts as covered, so that \(u\le\tau_q\) holds exactly when \(y\le\hat q_{\tau_q}\). Figure~\ref{fig:overview} and the case study use IR as a fixed method, chosen before seeing any test data. For the prediction intervals in Figure~\ref{fig:case_study}(a), IR is fitted on the extended grid of Appendix~\ref{app:robustness}, and the recalibrated 90\% interval spans the predicted quantiles at the levels whose calibration coverage is 5\% and 95\% (0.021 and 0.975), interpolated linearly between grid levels; the example forecast is the test window with the largest coverage gain.
	
	\newpage
	\section{Additional Results}
	\label{app:additional_results}
	
	\subsection{Additional calibration errors}
	Table~\ref{tab:scorecard_se} extends Table~\ref{tab:scorecard} with standard errors across the 55 dataset configurations. Because all models are evaluated on the same datasets, comparisons between models should rely on paired differences across datasets rather than on these standard errors.
	
	\begin{table*}[htb]
		\centering
		\scriptsize
		\setlength{\tabcolsep}{4pt}
		\renewcommand{\arraystretch}{1.15}
		\begin{tabular}{@{} c l c c c c c c c @{}}
			\toprule
			& & \multicolumn{3}{c}{\textbf{PCE}} & \multicolumn{2}{c}{\textbf{QCE}} & \multicolumn{2}{c}{\textbf{Diag.}} \\
			\cmidrule(lr){3-5} \cmidrule(lr){6-7} \cmidrule(lr){8-9}
			\textbf{Cat.} & \textbf{Model} & \textbf{pool} & \textbf{avg} & \textbf{max} & \textbf{avg} & \textbf{max} & \( \mathbf{\Delta_H} \) & \( \mathbf{\Delta_C} \) \\
			\midrule
			1 & \cellcolor{cat1color!12}\textbf{Moirai-Base} & \shortstack{0.0536\\{\color{gray}\(\pm 0.0073\)}} & \shortstack{0.0829\\{\color{gray}\(\pm 0.0085\)}} & \shortstack{0.1579\\{\color{gray}\(\pm 0.0142\)}} & \shortstack{0.1215\\{\color{gray}\(\pm 0.0100\)}} & \shortstack{0.1862\\{\color{gray}\(\pm 0.0144\)}} & \shortstack{0.0293\\{\color{gray}\(\pm 0.0046\)}} & \shortstack{0.0386\\{\color{gray}\(\pm 0.0040\)}} \\
			1 & \cellcolor{cat1color!12}\textbf{Moirai-Large} & \shortstack{0.0484\\{\color{gray}\(\pm 0.0065\)}} & \shortstack{0.0776\\{\color{gray}\(\pm 0.0080\)}} & \shortstack{0.1581\\{\color{gray}\(\pm 0.0149\)}} & \shortstack{0.1225\\{\color{gray}\(\pm 0.0099\)}} & \shortstack{0.1906\\{\color{gray}\(\pm 0.0151\)}} & \shortstack{0.0292\\{\color{gray}\(\pm 0.0040\)}} & \shortstack{0.0448\\{\color{gray}\(\pm 0.0055\)}} \\
			1 & \cellcolor{cat1color!12}\textbf{Moirai-MoE-Base} & \shortstack{0.0533\\{\color{gray}\(\pm 0.0066\)}} & \shortstack{0.0826\\{\color{gray}\(\pm 0.0082\)}} & \shortstack{0.1537\\{\color{gray}\(\pm 0.0145\)}} & \shortstack{0.1204\\{\color{gray}\(\pm 0.0098\)}} & \shortstack{0.1878\\{\color{gray}\(\pm 0.0157\)}} & \shortstack{0.0293\\{\color{gray}\(\pm 0.0041\)}} & \shortstack{0.0379\\{\color{gray}\(\pm 0.0042\)}} \\
			1 & \cellcolor{cat1color!12}\textbf{Moirai-MoE-Small} & \shortstack{0.0521\\{\color{gray}\(\pm 0.0067\)}} & \shortstack{0.0837\\{\color{gray}\(\pm 0.0082\)}} & \shortstack{0.1578\\{\color{gray}\(\pm 0.0136\)}} & \shortstack{0.1234\\{\color{gray}\(\pm 0.0097\)}} & \shortstack{0.1908\\{\color{gray}\(\pm 0.0148\)}} & \shortstack{0.0317\\{\color{gray}\(\pm 0.0044\)}} & \shortstack{0.0397\\{\color{gray}\(\pm 0.0047\)}} \\
			1 & \cellcolor{cat1color!12}\textbf{Moirai-Small} & \shortstack{0.0530\\{\color{gray}\(\pm 0.0070\)}} & \shortstack{0.0855\\{\color{gray}\(\pm 0.0092\)}} & \shortstack{0.1607\\{\color{gray}\(\pm 0.0146\)}} & \shortstack{0.1283\\{\color{gray}\(\pm 0.0101\)}} & \shortstack{0.1982\\{\color{gray}\(\pm 0.0146\)}} & \shortstack{0.0326\\{\color{gray}\(\pm 0.0049\)}} & \shortstack{0.0428\\{\color{gray}\(\pm 0.0049\)}} \\
			\midrule
			2 & \cellcolor{cat2color!12}\textbf{Chronos-T5-Base} & \shortstack{0.1024\\{\color{gray}\(\pm 0.0109\)}} & \shortstack{0.1240\\{\color{gray}\(\pm 0.0117\)}} & \shortstack{0.1939\\{\color{gray}\(\pm 0.0174\)}} & \shortstack{0.1491\\{\color{gray}\(\pm 0.0124\)}} & \shortstack{0.2159\\{\color{gray}\(\pm 0.0177\)}} & \shortstack{0.0216\\{\color{gray}\(\pm 0.0041\)}} & \shortstack{0.0251\\{\color{gray}\(\pm 0.0037\)}} \\
			2 & \cellcolor{cat2color!12}\textbf{Chronos-T5-Small} & \shortstack{0.1035\\{\color{gray}\(\pm 0.0112\)}} & \shortstack{0.1253\\{\color{gray}\(\pm 0.0115\)}} & \shortstack{0.1960\\{\color{gray}\(\pm 0.0171\)}} & \shortstack{0.1524\\{\color{gray}\(\pm 0.0122\)}} & \shortstack{0.2183\\{\color{gray}\(\pm 0.0173\)}} & \shortstack{0.0218\\{\color{gray}\(\pm 0.0046\)}} & \shortstack{0.0271\\{\color{gray}\(\pm 0.0041\)}} \\
			2 & \cellcolor{cat2color!12}\textbf{TabPFN-TS} & \shortstack{0.0629\\{\color{gray}\(\pm 0.0062\)}} & \shortstack{0.0872\\{\color{gray}\(\pm 0.0079\)}} & \shortstack{0.1537\\{\color{gray}\(\pm 0.0137\)}} & \shortstack{0.1198\\{\color{gray}\(\pm 0.0094\)}} & \shortstack{0.1797\\{\color{gray}\(\pm 0.0135\)}} & \shortstack{0.0243\\{\color{gray}\(\pm 0.0045\)}} & \shortstack{0.0325\\{\color{gray}\(\pm 0.0036\)}} \\
			\midrule
			3 & \cellcolor{cat3color!12}\textbf{Chronos-2} & \shortstack{0.0641\\{\color{gray}\(\pm 0.0082\)}} & \shortstack{0.0919\\{\color{gray}\(\pm 0.0094\)}} & \shortstack{0.1653\\{\color{gray}\(\pm 0.0147\)}} & \shortstack{0.1229\\{\color{gray}\(\pm 0.0107\)}} & \shortstack{0.1881\\{\color{gray}\(\pm 0.0150\)}} & \shortstack{0.0278\\{\color{gray}\(\pm 0.0043\)}} & \shortstack{0.0310\\{\color{gray}\(\pm 0.0040\)}} \\
			3 & \cellcolor{cat3color!12}\textbf{Chronos-Bolt-Base} & \shortstack{0.0505\\{\color{gray}\(\pm 0.0063\)}} & \shortstack{0.0804\\{\color{gray}\(\pm 0.0086\)}} & \shortstack{0.1494\\{\color{gray}\(\pm 0.0147\)}} & \shortstack{0.1202\\{\color{gray}\(\pm 0.0104\)}} & \shortstack{0.1838\\{\color{gray}\(\pm 0.0150\)}} & \shortstack{0.0299\\{\color{gray}\(\pm 0.0044\)}} & \shortstack{0.0398\\{\color{gray}\(\pm 0.0048\)}} \\
			3 & \cellcolor{cat3color!12}\textbf{Chronos-Bolt-Small} & \shortstack{0.0523\\{\color{gray}\(\pm 0.0053\)}} & \shortstack{0.0833\\{\color{gray}\(\pm 0.0078\)}} & \shortstack{0.1569\\{\color{gray}\(\pm 0.0135\)}} & \shortstack{0.1225\\{\color{gray}\(\pm 0.0098\)}} & \shortstack{0.1890\\{\color{gray}\(\pm 0.0142\)}} & \shortstack{0.0310\\{\color{gray}\(\pm 0.0042\)}} & \shortstack{0.0393\\{\color{gray}\(\pm 0.0045\)}} \\
			3 & \cellcolor{cat3color!12}\textbf{TiRex} & \shortstack{0.0502\\{\color{gray}\(\pm 0.0068\)}} & \shortstack{0.0796\\{\color{gray}\(\pm 0.0084\)}} & \shortstack{0.1485\\{\color{gray}\(\pm 0.0139\)}} & \shortstack{0.1161\\{\color{gray}\(\pm 0.0102\)}} & \shortstack{0.1798\\{\color{gray}\(\pm 0.0150\)}} & \shortstack{0.0294\\{\color{gray}\(\pm 0.0041\)}} & \shortstack{0.0365\\{\color{gray}\(\pm 0.0048\)}} \\
			3 & \cellcolor{cat3color!12}\textbf{TimesFM-2.5-200M} & \shortstack{0.0743\\{\color{gray}\(\pm 0.0105\)}} & \shortstack{0.1011\\{\color{gray}\(\pm 0.0109\)}} & \shortstack{0.1711\\{\color{gray}\(\pm 0.0156\)}} & \shortstack{0.1254\\{\color{gray}\(\pm 0.0113\)}} & \shortstack{0.1900\\{\color{gray}\(\pm 0.0158\)}} & \shortstack{0.0268\\{\color{gray}\(\pm 0.0041\)}} & \shortstack{0.0243\\{\color{gray}\(\pm 0.0037\)}} \\
			\midrule
			4 & \cellcolor{cat4color!12}\textbf{Sundial-Base} & \shortstack{0.1467\\{\color{gray}\(\pm 0.0072\)}} & \shortstack{0.1678\\{\color{gray}\(\pm 0.0092\)}} & \shortstack{0.2413\\{\color{gray}\(\pm 0.0137\)}} & \shortstack{0.1958\\{\color{gray}\(\pm 0.0107\)}} & \shortstack{0.2629\\{\color{gray}\(\pm 0.0143\)}} & \shortstack{0.0211\\{\color{gray}\(\pm 0.0047\)}} & \shortstack{0.0280\\{\color{gray}\(\pm 0.0049\)}} \\
			\bottomrule
		\end{tabular}
		\caption{Calibration metrics across probabilistic TSFMs with standard errors across datasets (mean on the first line, standard error on the second). Lower is better for all calibration errors and diagnostics.}
		\label{tab:scorecard_se}
	\end{table*}
	
\begin{figure}[htb]
		\centering
		\includegraphics[width=1\linewidth]{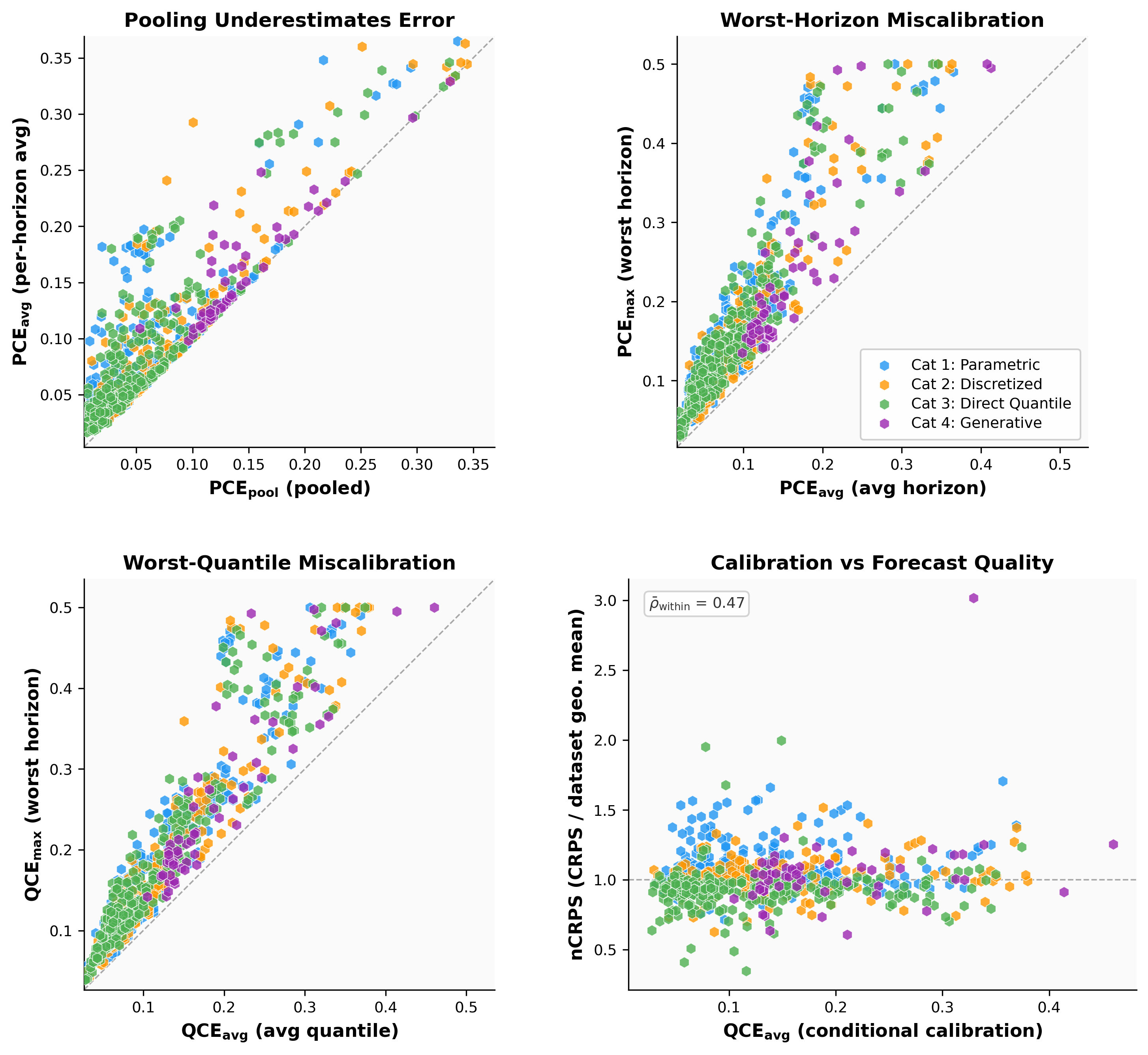}
		\caption{Comparison of aggregate and disaggregated calibration diagnostics for all model--dataset pairs. \textbf{Top left:} \( \mathrm{PCE}_{\mathrm{pool}} \) versus \( \mathrm{PCE}_{\mathrm{avg}} \); pooling across horizons understates the horizon-wise error. \textbf{Top right:} \( \mathrm{PCE}_{\mathrm{avg}} \) versus \( \mathrm{PCE}_{\max} \); moderate average miscalibration can coincide with severe failures at individual horizons. \textbf{Bottom left:} \( \mathrm{QCE}_{\mathrm{avg}} \) versus \( \mathrm{QCE}_{\max} \). \textbf{Bottom right:} \( \mathrm{QCE}_{\mathrm{avg}} \) versus relative CRPS (CRPS divided by its geometric mean over models on the same dataset); better-calibrated models tend to be more accurate, but calibration and accuracy are clearly distinct properties. Colors denote uncertainty representation categories. Because the maxima are taken over horizons, \( \mathrm{PCE}_{\max} \) and \( \mathrm{QCE}_{\max} \) grow with the horizon length and are best compared within a dataset.}
		\label{fig:add_cal_errors}
	\end{figure}

	Figure~\ref{fig:add_cal_errors} shows, for each model--dataset pair, how aggregation choices affect the apparent severity of miscalibration and how calibration relates to forecast accuracy.
	
	\subsection{Further case studies}
	\label{app:case_studies}
	
	\paragraph{Value dependence across datasets.}
	Table~\ref{tab:case_dc} lists further model--dataset pairs with small probabilistic calibration errors but significant conditioning gaps. For each, it reports how often outcomes fall below the predicted 10\% quantile and above the predicted 90\% quantile when the respective predicted quantile is in its lowest or highest within-horizon quintile (nominal: 10\% everywhere). The pattern of the main-text case study recurs in every row: outcomes fall below the lower quantile far more often when the model predicts high values, and above the upper quantile more often when it predicts low values. The forecasts are therefore too extreme in both directions, and these errors cancel in marginal coverage. The last row is the intermittent-demand dataset \texttt{car\_parts}, where 79\% of the targets are zero. Its very large conditioning gap is genuine in the sense that the model's lower quantiles are negative for low-demand items and never covered, but much of it reflects the point mass at zero (Appendix~\ref{sec:calib_computation}), and it should not be compared directly with the other rows.
	
	\begin{table}[htb]
		\centering
		\small
		\setlength{\tabcolsep}{3.5pt}
		\resizebox{\textwidth}{!}{%
			\begin{tabular}{@{} l l rrrr rr rr r @{}} \toprule
				& & & & & & \multicolumn{2}{c}{\textbf{below \(\hat q_{0.1}\)}} & \multicolumn{2}{c}{\textbf{above \(\hat q_{0.9}\)}} & \\
				\cmidrule(lr){7-8} \cmidrule(lr){9-10}
				\textbf{Model} & \textbf{Dataset} & \(\mathrm{PCE}_{\mathrm{pool}}\) & \(\mathrm{PCE}_{\mathrm{avg}}\) & \(\mathrm{QCE}_{\mathrm{avg}}\) & \(\Delta_C\) corr. & \textbf{lowest} & \textbf{highest} & \textbf{lowest} & \textbf{highest} & \textbf{point mass} \\
				\midrule
				\cellcolor{cat2color!12}Chronos-T5-Small & \texttt{LOOP\_SEATTLE/5T} & 0.031 & 0.034 & 0.079 & 0.030 & 15\% & 22\% & 29\% & 4\% & 0\% \\
				\cellcolor{cat1color!12}Moirai-Small & \texttt{kdd\_cup\_2018/H} & 0.015 & 0.032 & 0.071 & 0.028 & 2\% & 26\% & 12\% & 9\% & 2\% \\
				\cellcolor{cat1color!12}Moirai-Base & \texttt{electricity/D} & 0.023 & 0.034 & 0.075 & 0.031 & 8\% & 17\% & 14\% & 5\% & 0\% \\
				\cellcolor{cat3color!12}TiRex & \texttt{electricity/W} & 0.009 & 0.039 & 0.070 & 0.020 & 6\% & 12\% & 11\% & 4\% & 0\% \\
				\cellcolor{cat2color!12}TabPFN-TS & \texttt{m4\_hourly} & 0.036 & 0.050 & 0.088 & 0.030 & 6\% & 15\% & 17\% & 10\% & 1\% \\
				\cellcolor{cat1color!12}Moirai-MoE-Base & \texttt{m4\_quarterly} & 0.016 & 0.022 & 0.058 & 0.025 & 6\% & 12\% & 12\% & 6\% & 0\% \\
				\cellcolor{cat1color!12}Moirai-Large & \texttt{car\_parts} & 0.038 & 0.050 & 0.239 & 0.182 & 0\% & 37\% & 12\% & 3\% & 79\% \\
				\bottomrule
		\end{tabular}}
		\caption{Model--dataset pairs with value-dependent miscalibration. ``lowest''/``highest'': share of outcomes below \(\hat q_{0.1}\) (or above \(\hat q_{0.9}\)) among forecasts whose predicted quantile lies in the lowest or highest within-horizon quintile; the nominal value is 10\%. ``point mass'': share of targets equal to the most frequent value of the dataset. All conditioning gaps are significant (\(p\le0.05\)).}
		\label{tab:case_dc}
	\end{table}
	
\begin{figure}[htb]
		\centering
		\includegraphics[width=\linewidth]{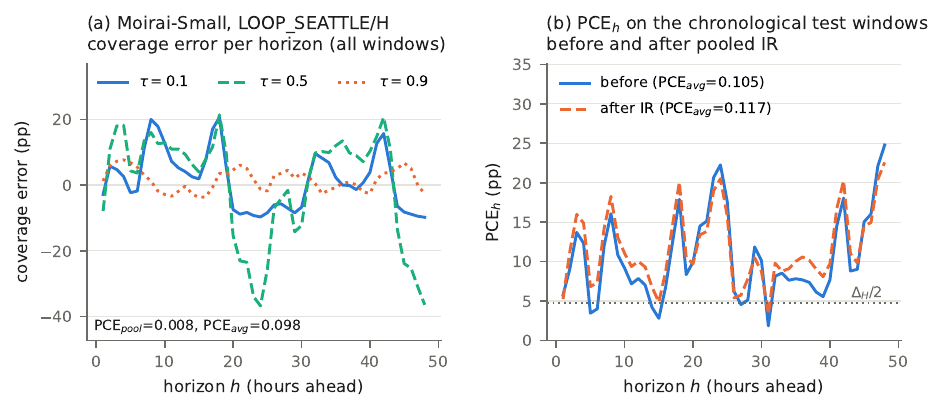}
		\caption{Horizon cancellation for Moirai-Small on \texttt{LOOP\_SEATTLE/H}. \textbf{(a)} Coverage error \(\hat p_h(\tau)-\tau\) per horizon for three levels. \textbf{(b)} \(\mathrm{PCE}_h\) on the later 80\% of the windows before and after isotonic recalibration fitted on the earliest 20\%, with the lower bound \(\Delta_H/2\) of Theorem~\ref{thm:global_relabelling}(ii) computed on the same windows.}
		\label{fig:case_dh}
	\end{figure}

	\paragraph{Horizon cancellation.}
	Figure~\ref{fig:case_dh} shows the complementary failure mode. Moirai-Small on hourly traffic speeds (\texttt{LOOP\_SEATTLE/H}) appears almost perfectly calibrated under pooled coverage (\(\mathrm{PCE}_{\mathrm{pool}}=0.008\)). Its coverage, however, oscillates with the time of day being forecast, with errors of up to 38 percentage points that cancel across the 48 horizons (\(\mathrm{PCE}_{\mathrm{avg}}=0.098\), \(\Delta_H=0.089\)). A practitioner who uses the intervals for a particular hour of the next day would face these errors in full. Global recalibration fitted on the earliest windows cannot remove them and even increases the horizon-wise error on the later windows (0.105\(\,\to\,\)0.117), which stays above the floor \(\Delta_H/2\) of Theorem~\ref{thm:global_relabelling}(ii).
	
	\subsection{Additional recalibration results}
	\label{app:additional_recal}
	
	\paragraph{Worst-horizon errors and gaps.}
	Table~\ref{tab:recalibration_additional} reports the relative change of the worst-horizon errors and of both gaps under the chronological split, computed as in Table~\ref{tab:recalibration}. Worst-horizon errors improve much less than their averages. The horizon gap increases for every model and method: recalibration removes the pooled error, which is the part that \(\Delta_H\) does not measure, while the cancellation across horizons remains (Theorem~\ref{thm:global_relabelling}(ii)). The conditioning gap changes little on average, so in our experiments global relabeling rarely creates or removes value-dependent miscalibration. Its relative increase for Chronos-T5 and Sundial reflects that their raw conditioning gaps are small; once their large marginal errors are removed, the value-dependent part becomes more prominent.
	
	\begin{table*}[htb]
		\centering
		\small
		\setlength{\tabcolsep}{4pt}
		\renewcommand{\arraystretch}{1.12}
		\resizebox{\textwidth}{!}{%
			\begin{tabular}{@{} l rrr rrr rrr rrr @{}} \toprule
				& \multicolumn{3}{c}{\textbf{\( \mathbf{PCE}_{\mathbf{max}} \)}}
				& \multicolumn{3}{c}{\textbf{\( \mathbf{QCE}_{\mathbf{max}} \)}}
				& \multicolumn{3}{c}{\textbf{\( \mathbf{\Delta_H} \)}}
				& \multicolumn{3}{c}{\textbf{\( \mathbf{\Delta_C} \)}} \\
				\cmidrule(lr){2-4} \cmidrule(lr){5-7} \cmidrule(lr){8-10} \cmidrule(l){11-13}
				\textbf{Model} & \textbf{IR} & \textbf{KDE} & \textbf{PS} & \textbf{IR} & \textbf{KDE} & \textbf{PS} & \textbf{IR} & \textbf{KDE} & \textbf{PS} & \textbf{IR} & \textbf{KDE} & \textbf{PS} \\
				\midrule
				\cellcolor{cat1color!12}\textbf{Moirai-Base} & $-$1.67\% & $-$1.53\% & \textbf{$-$1.98\%} & $-$1.80\% & $-$1.75\% & \textbf{$-$2.31\%} & $+$9.86\% & \textbf{$+$9.64\%} & $+$9.69\% & $-$8.46\% & \textbf{$-$8.48\%} & $-$7.22\% \\
				\cellcolor{cat1color!12}\textbf{Moirai-Large} & \textbf{$-$0.86\%} & $-$0.73\% & $+$0.40\% & \textbf{$-$0.76\%} & \textbf{$-$0.76\%} & $+$0.02\% & $+$17.00\% & $+$16.62\% & \textbf{$+$15.70\%} & $-$3.71\% & \textbf{$-$3.87\%} & $-$3.44\% \\
				\cellcolor{cat1color!12}\textbf{Moirai-MoE-Base} & \textbf{$-$3.43\%} & $-$3.37\% & $-$3.06\% & $-$3.00\% & \textbf{$-$3.06\%} & $-$2.59\% & $+$16.67\% & $+$16.68\% & \textbf{$+$11.82\%} & \textbf{$-$4.68\%} & $-$4.54\% & $-$4.51\% \\
				\cellcolor{cat1color!12}\textbf{Moirai-MoE-Small} & $+$0.33\% & $+$0.31\% & \textbf{$+$0.20\%} & \textbf{$-$0.63\%} & $-$0.58\% & $-$0.58\% & $+$7.00\% & $+$7.30\% & \textbf{$+$5.27\%} & \textbf{$-$0.19\%} & $+$0.03\% & $+$0.50\% \\
				\cellcolor{cat1color!12}\textbf{Moirai-Small} & $+$2.64\% & $+$2.78\% & \textbf{$+$2.13\%} & $+$1.46\% & $+$1.44\% & \textbf{$+$1.04\%} & \textbf{$+$4.69\%} & $+$5.18\% & $+$5.72\% & $+$0.93\% & $+$0.38\% & \textbf{$+$0.11\%} \\
				\midrule
				\cellcolor{cat2color!12}\textbf{Chronos-T5-Base} & \textbf{$-$17.59\%} & $-$17.43\% & $-$16.97\% & \textbf{$-$15.23\%} & $-$15.03\% & $-$15.02\% & $+$66.79\% & $+$66.71\% & \textbf{$+$64.09\%} & $+$26.71\% & $+$27.12\% & \textbf{$+$24.91\%} \\
				\cellcolor{cat2color!12}\textbf{Chronos-T5-Small} & $-$16.43\% & $-$16.42\% & \textbf{$-$16.69\%} & $-$14.72\% & $-$14.67\% & \textbf{$-$14.93\%} & $+$55.23\% & $+$55.10\% & \textbf{$+$51.71\%} & $+$26.90\% & $+$27.23\% & \textbf{$+$26.33\%} \\
				\cellcolor{cat2color!12}\textbf{TabPFN-TS} & $-$3.70\% & \textbf{$-$3.78\%} & $-$2.74\% & $-$6.32\% & \textbf{$-$6.36\%} & $-$5.81\% & $+$9.43\% & $+$9.73\% & \textbf{$+$4.85\%} & $-$7.80\% & $-$7.72\% & \textbf{$-$8.05\%} \\
				\midrule
				\cellcolor{cat3color!12}\textbf{Chronos-2} & $-$5.86\% & $-$5.68\% & \textbf{$-$6.32\%} & $-$6.51\% & $-$6.47\% & \textbf{$-$6.83\%} & $+$35.86\% & $+$36.52\% & \textbf{$+$32.24\%} & \textbf{$-$3.57\%} & $-$3.42\% & $+$0.86\% \\
				\cellcolor{cat3color!12}\textbf{Chronos-Bolt-Base} & \textbf{$+$0.81\%} & $+$0.96\% & $+$1.80\% & $-$0.40\% & \textbf{$-$0.42\%} & $+$0.08\% & $+$10.03\% & $+$10.27\% & \textbf{$+$5.42\%} & $-$6.29\% & $-$6.74\% & \textbf{$-$6.75\%} \\
				\cellcolor{cat3color!12}\textbf{Chronos-Bolt-Small} & \textbf{$-$4.80\%} & $-$4.51\% & $-$3.74\% & \textbf{$-$4.53\%} & $-$4.45\% & $-$3.71\% & $+$18.02\% & $+$18.10\% & \textbf{$+$12.04\%} & $-$0.77\% & \textbf{$-$1.27\%} & $-$0.54\% \\
				\cellcolor{cat3color!12}\textbf{TiRex} & $+$1.11\% & $+$1.20\% & \textbf{$+$0.90\%} & $-$2.62\% & \textbf{$-$2.71\%} & $-$1.75\% & $+$15.00\% & $+$14.70\% & \textbf{$+$11.04\%} & $-$10.95\% & \textbf{$-$11.30\%} & $-$8.73\% \\
				\cellcolor{cat3color!12}\textbf{TimesFM-2.5-200M} & \textbf{$-$11.45\%} & $-$11.38\% & $-$10.69\% & \textbf{$-$11.05\%} & $-$10.98\% & $-$10.32\% & $+$36.86\% & $+$36.81\% & \textbf{$+$29.80\%} & $-$4.47\% & $-$4.43\% & \textbf{$-$4.97\%} \\
				\midrule
				\cellcolor{cat4color!12}\textbf{Sundial-Base} & $-$14.19\% & $-$14.19\% & \textbf{$-$14.43\%} & $-$14.38\% & $-$14.37\% & \textbf{$-$14.46\%} & $+$66.25\% & $+$66.37\% & \textbf{$+$65.29\%} & $+$13.74\% & $+$13.61\% & \textbf{$+$13.43\%} \\
				\bottomrule
			\end{tabular}%
		}
		\caption{Relative change of worst-horizon errors and diagnostic gaps under the chronological split, computed as in Table~\ref{tab:recalibration}. Negative values are decreases, positive values increases.}
		\label{tab:recalibration_additional}
	\end{table*}
	
	\paragraph{Random split.}
	Table~\ref{tab:recalibration_random} repeats Table~\ref{tab:recalibration} for a random split of the forecast windows. The gains are larger (IR: 67\% for \(\mathrm{PCE}_{\mathrm{pool}}\), 35\% for \(\mathrm{PCE}_{\mathrm{avg}}\), and 24\% for \(\mathrm{QCE}_{\mathrm{avg}}\), with 80\% of the pairs improved) likely because calibration and test windows are interleaved in time, so the calibration data represent the test period better. The ordering across metrics and the near-equivalence of the methods are the same as under the chronological split.
	
	\begin{table*}[htb]
		\centering
		\small
		\setlength{\tabcolsep}{4.5pt}
		\renewcommand{\arraystretch}{1.12}
		\resizebox{\textwidth}{!}{%
			\begin{tabular}{@{} l rrr rrr rrr @{}} \toprule
				& \multicolumn{3}{c}{\textbf{\( \mathbf{PCE}_{\mathbf{pool}} \)}}
				& \multicolumn{3}{c}{\textbf{\( \mathbf{PCE}_{\mathbf{avg}} \)}}
				& \multicolumn{3}{c}{\textbf{\( \mathbf{QCE}_{\mathbf{avg}} \)}} \\
				\cmidrule(lr){2-4} \cmidrule(lr){5-7} \cmidrule(l){8-10}
				\textbf{Model} & \textbf{IR} & \textbf{KDE} & \textbf{PS} & \textbf{IR} & \textbf{KDE} & \textbf{PS} & \textbf{IR} & \textbf{KDE} & \textbf{PS} \\
				\midrule
				\cellcolor{cat1color!12}\textbf{Moirai-Base} & \textbf{$-$53.45\%} & $-$53.23\% & $-$46.30\% & \textbf{$-$22.56\%} & $-$22.39\% & $-$20.86\% & \textbf{$-$14.60\%} & $-$14.47\% & $-$13.40\% \\
				\cellcolor{cat1color!12}\textbf{Moirai-Large} & $-$44.36\% & \textbf{$-$44.47\%} & $-$39.68\% & \textbf{$-$14.84\%} & $-$14.72\% & $-$13.80\% & \textbf{$-$9.53\%} & $-$9.45\% & $-$8.76\% \\
				\cellcolor{cat1color!12}\textbf{Moirai-MoE-Base} & \textbf{$-$57.14\%} & $-$56.55\% & $-$48.77\% & \textbf{$-$22.02\%} & $-$21.82\% & $-$19.81\% & \textbf{$-$14.50\%} & $-$14.30\% & $-$12.83\% \\
				\cellcolor{cat1color!12}\textbf{Moirai-MoE-Small} & \textbf{$-$59.97\%} & $-$59.06\% & $-$48.86\% & \textbf{$-$22.22\%} & $-$22.03\% & $-$18.87\% & \textbf{$-$13.64\%} & $-$13.48\% & $-$11.85\% \\
				\cellcolor{cat1color!12}\textbf{Moirai-Small} & \textbf{$-$47.60\%} & $-$47.43\% & $-$38.48\% & \textbf{$-$17.46\%} & $-$17.41\% & $-$14.40\% & $-$9.14\% & \textbf{$-$9.16\%} & $-$7.41\% \\
				\midrule
				\cellcolor{cat2color!12}\textbf{Chronos-T5-Base} & $-$78.28\% & \textbf{$-$78.52\%} & $-$74.57\% & $-$51.31\% & \textbf{$-$51.34\%} & $-$48.85\% & \textbf{$-$40.97\%} & $-$40.96\% & $-$38.80\% \\
				\cellcolor{cat2color!12}\textbf{Chronos-T5-Small} & $-$78.34\% & \textbf{$-$78.57\%} & $-$69.57\% & $-$51.26\% & \textbf{$-$51.32\%} & $-$47.79\% & \textbf{$-$40.00\%} & \textbf{$-$40.00\%} & $-$36.71\% \\
				\cellcolor{cat2color!12}\textbf{TabPFN-TS} & \textbf{$-$60.14\%} & $-$59.95\% & $-$53.30\% & \textbf{$-$30.11\%} & $-$29.92\% & $-$28.97\% & \textbf{$-$20.67\%} & $-$20.49\% & $-$19.99\% \\
				\midrule
				\cellcolor{cat3color!12}\textbf{Chronos-2} & \textbf{$-$69.14\%} & $-$68.65\% & $-$63.09\% & \textbf{$-$33.86\%} & $-$33.70\% & $-$32.57\% & \textbf{$-$25.35\%} & $-$25.23\% & $-$24.36\% \\
				\cellcolor{cat3color!12}\textbf{Chronos-Bolt-Base} & $-$52.48\% & \textbf{$-$52.53\%} & $-$48.73\% & \textbf{$-$21.13\%} & $-$21.03\% & $-$20.55\% & $-$11.46\% & \textbf{$-$11.47\%} & $-$11.03\% \\
				\cellcolor{cat3color!12}\textbf{Chronos-Bolt-Small} & \textbf{$-$56.14\%} & $-$55.51\% & $-$42.30\% & \textbf{$-$22.91\%} & $-$22.76\% & $-$18.28\% & \textbf{$-$13.77\%} & $-$13.72\% & $-$11.08\% \\
				\cellcolor{cat3color!12}\textbf{TiRex} & \textbf{$-$57.62\%} & $-$57.32\% & $-$53.61\% & \textbf{$-$23.23\%} & $-$23.00\% & $-$22.70\% & \textbf{$-$16.55\%} & $-$16.51\% & $-$16.07\% \\
				\cellcolor{cat3color!12}\textbf{TimesFM-2.5-200M} & \textbf{$-$75.56\%} & $-$75.38\% & $-$68.65\% & \textbf{$-$44.49\%} & $-$44.41\% & $-$41.12\% & \textbf{$-$37.66\%} & $-$37.57\% & $-$34.72\% \\
				\midrule
				\cellcolor{cat4color!12}\textbf{Sundial-Base} & \textbf{$-$82.96\%} & \textbf{$-$82.96\%} & $-$82.12\% & \textbf{$-$57.77\%} & $-$57.75\% & $-$57.48\% & \textbf{$-$45.64\%} & $-$45.63\% & $-$45.47\% \\
				\bottomrule
			\end{tabular}%
		}
		\caption{Relative change of the calibration errors under a random 20/80 split of the forecast windows; layout as in Table~\ref{tab:recalibration}.}
		\label{tab:recalibration_random}
	\end{table*}
	
\begin{figure}[!htb]
		\centering
		\includegraphics[width=0.9\linewidth]{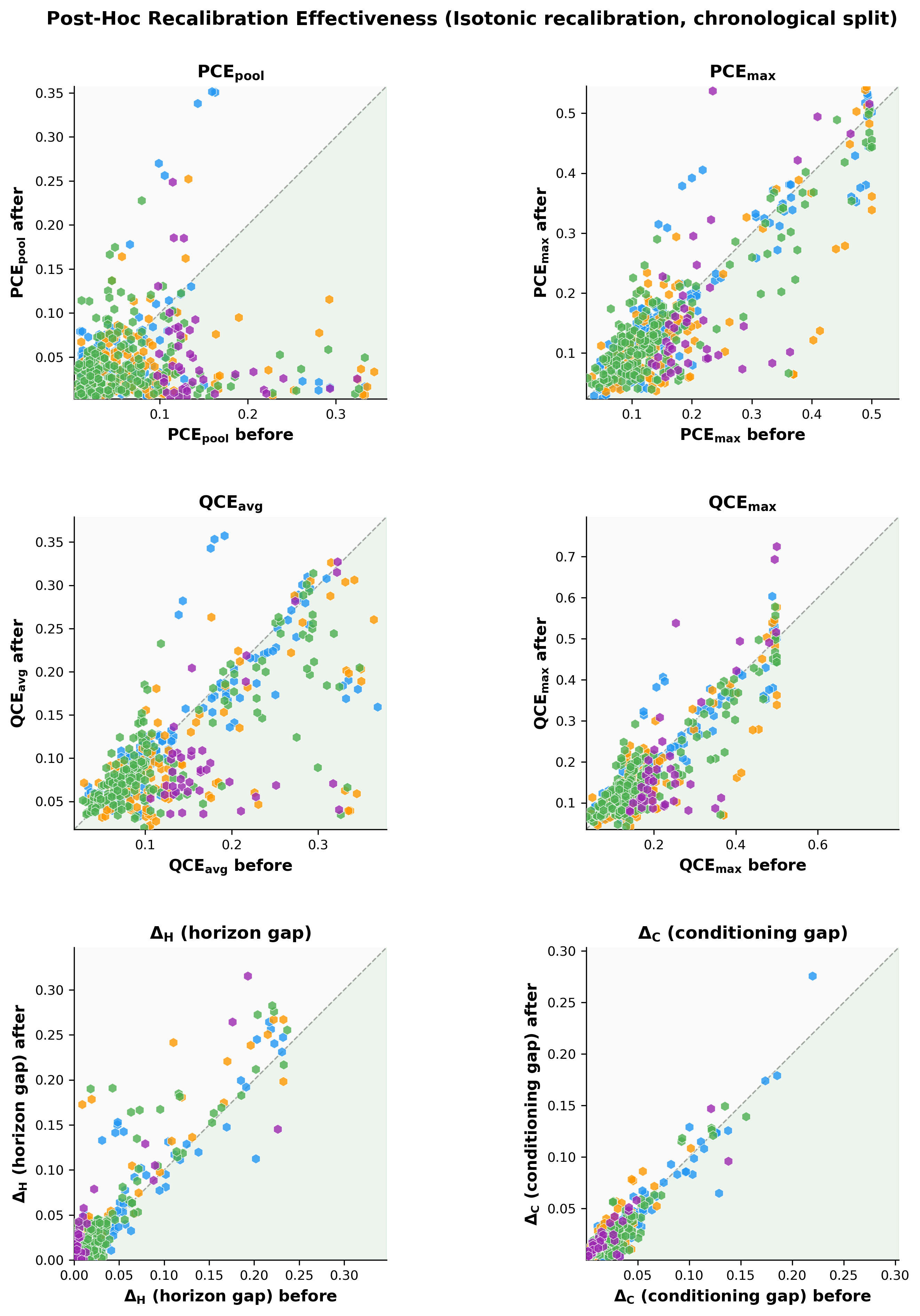}
		\caption{Effect of isotonic recalibration on each model--dataset pair under the chronological split. Each panel compares an error before (x-axis) and after (y-axis) recalibration; points below the diagonal improve. Gains concentrate in \( \mathrm{PCE}_{\mathrm{pool}} \) and in strongly miscalibrated pairs. Worst-horizon errors and the gaps \( \Delta_H \) and \( \Delta_C \) improve much less often, and \( \Delta_H \) typically increases.}
		\label{fig:add_recalibration}
	\end{figure}

	\paragraph{Per-pair effects and method selection.}
	Figure~\ref{fig:add_recalibration} shows the effect of IR on every model--dataset pair. Selecting for each pair the method with the lowest test error, which is not possible in deployment, improves 63\% of the pairs under the chronological split (82\% under the random split), compared with 62\% (80\%) for the fixed IR map. A fixed method therefore loses little.
	
	\subsection{Relabeling maps on a quantile grid}
	\label{app:relabeling}
	
	All four recalibration methods learn a single non-decreasing map from nominal to recalibrated levels. They differ in what they use from the calibration data (coverages at the grid levels for IR and PS, the distribution of the PIT values for HB and KDE) and in how they behave between and beyond the fitted levels. For quantile-grid forecasts, these differences shrink at the fitted levels: IR and HB coincide there, while KDE and PS smooth or approximate the same empirical coverage curve.
	
	\begin{proposition}
		\label{prop:grid_equivalence}
		Let \(\hat p(\tau_q)\) be the pooled empirical coverage of the \(\tau_q\)-quantiles on the calibration windows and \(\hat F_U\) the empirical CDF of the approximate PIT values of Appendix~\ref{sec:recalibration}. Then \(\hat p\) is non-decreasing in \(q\), the IR map satisfies \(r_{\mathrm{IR}}(\tau_q)=\hat p(\tau_q)\), and \(\hat F_U(\tau_q)=\hat p(\tau_q)\). Consequently, HB with bin edges that contain the grid satisfies \(r_{\mathrm{HB}}(\tau_q)=r_{\mathrm{IR}}(\tau_q)\) at every grid level, KDE returns a smoothed version of \(\hat F_U(\tau_q)\), and PS returns the best logit-affine approximation of \(\hat p\).
	\end{proposition}
	
	\begin{proof}
		The predicted quantiles of each forecast are ordered, so a forecast that covers at level \(\tau_q\) also covers at every higher level, and \(\hat p\) is non-decreasing; isotonic regression leaves a non-decreasing sequence unchanged. By construction of the approximate PIT, \(u\le\tau_q\) holds if and only if \(y\le\hat q_{\tau_q}\), and averaging gives \(\hat F_U(\tau_q)=\hat p(\tau_q)\). HB evaluates \(\hat F_U\) at its bin edges, and the statements for KDE and PS follow from their definitions.
	\end{proof}
	
	In our implementation, IR is evaluated through interpolation on a 1000-point grid, so HB and IR differ by at most 0.0001 in any error of any model--dataset pair; we therefore report HB only here. The similarity of the methods in the main experiments is therefore expected: at the fitted levels they estimate the same coverage curve, giving similar but not identical results, and the remaining errors are largely structural (Theorem~\ref{thm:global_relabelling}) rather than specific to one estimator. The methods do differ when recalibrated levels are needed off the fitted grid. We illustrate this with the extended-grid forecasts of Appendix~\ref{app:robustness} (four models, 42 configurations, chronological split). In design~A every map is fitted on all thirteen levels \(\{0.01,0.05,0.1,\dots,0.9,0.95,0.99\}\); in design~B it is fitted on \(\{0.1,\dots,0.9\}\) only and then applied to all thirteen, as when a model recalibrated on deciles is used for 1\% and 99\% quantiles (Table~\ref{tab:tails_recal}). In design~A, IR is exact at every fitted level and is best throughout. In design~B, IR is constant outside the fitted range, so the 1\% quantile receives the level of the 10\% quantile and its tail error is no better than without recalibration, whereas PS extrapolates in logit space and more than halves the tail error compared with no recalibration. In this comparison, IR is thus preferable when recalibrated levels are only needed at fitted levels, and PS when they must be extrapolated; neither choice changes the structural limits of global relabeling.
	
	\begin{table}[htb]
		\centering
		\small
		\setlength{\tabcolsep}{3.5pt}
		\begin{tabular}{@{} l c cccc cccc @{}} \toprule
			& & \multicolumn{4}{c}{\textbf{A: fitted on all 13 levels}} & \multicolumn{4}{c}{\textbf{B: fitted on 0.1--0.9}} \\
			\cmidrule(lr){3-6} \cmidrule(l){7-10}
			\(\tau\) & \textbf{None} & \textbf{IR} & \textbf{HB} & \textbf{KDE} & \textbf{PS} & \textbf{IR} & \textbf{HB} & \textbf{KDE} & \textbf{PS} \\
			\midrule
			0.01 & 0.077 & \textbf{0.019} & 0.061 & 0.029 & 0.031 & 0.103 & 0.060 & 0.066 & \textbf{0.039} \\
			0.05 & 0.098 & \textbf{0.030} & \textbf{0.030} & \textbf{0.030} & 0.034 & 0.058 & 0.058 & 0.057 & \textbf{0.041} \\
			0.1--0.9 & 0.091 & \textbf{0.047} & \textbf{0.047} & \textbf{0.047} & 0.050 & \textbf{0.047} & \textbf{0.047} & \textbf{0.047} & 0.049 \\
			0.95 & 0.078 & \textbf{0.029} & \textbf{0.029} & 0.031 & 0.032 & 0.053 & 0.053 & 0.053 & \textbf{0.034} \\
			0.99 & 0.053 & \textbf{0.017} & 0.041 & 0.021 & 0.022 & 0.106 & 0.037 & 0.064 & \textbf{0.027} \\
			\midrule
			\textit{Tails: vs.\ IR} & -- & -- & +0.016 & +0.004 & +0.006 & -- & -0.028 & -0.020 & -0.045 \\
			\bottomrule
		\end{tabular}
		\caption{Pooled coverage error \(|\bar p_{\mathrm{test}}(\tau)-r(\tau)|\) of the recalibrated levels on the extended grid, chronological split, mean over four models and 42 configurations. Bold marks the best method per design and row. The last row gives the mean difference to IR over the four tail levels. At the fitted levels the methods are similar; they separate off the fitted grid.}
		\label{tab:tails_recal}
	\end{table}
	
	\subsection{Robustness analyses}
	\label{app:robustness}
	
	\paragraph{Tail quantiles.}
	Table~\ref{tab:tails} reports per-level errors on the extended grid \(\{0.01,0.05,0.1,\dots,0.9,0.95,0.99\}\) for one model per category: quantiles of the three sample-based models are computed from \(S=500\) samples, and Chronos-2 was run with the extended grid. For the better-calibrated models (Moirai-Base, Chronos-2), the coverage errors shrink towards the tails. Chronos-T5 and Sundial remain badly miscalibrated in the tails: at \(\tau=0.01\) their coverage errors exceed ten times the nominal tail probability. Conclusions drawn from the central grid therefore carry over to the tails.
	
	\begin{table}[htb]
		\centering
		\small
		\setlength{\tabcolsep}{4pt}
		\begin{tabular}{@{} c cc cc cc cc @{}} \toprule
			& \multicolumn{2}{>{\columncolor{cat1color!12}}c}{\textbf{Moirai-Base}} & \multicolumn{2}{>{\columncolor{cat2color!12}}c}{\textbf{Chronos-T5-S}} & \multicolumn{2}{>{\columncolor{cat3color!12}}c}{\textbf{Chronos-2}} & \multicolumn{2}{>{\columncolor{cat4color!12}}c}{\textbf{Sundial}} \\
			\cmidrule(lr){2-3} \cmidrule(lr){4-5} \cmidrule(lr){6-7} \cmidrule(l){8-9}
			\(\tau\) & PCE\(_h\) & QCE\(_h\) & PCE\(_h\) & QCE\(_h\) & PCE\(_h\) & QCE\(_h\) & PCE\(_h\) & QCE\(_h\) \\
			\midrule
			0.01 & 0.010 & 0.015 & 0.133 & 0.135 & 0.010 & 0.014 & 0.154 & 0.156 \\
			0.05 & 0.031 & 0.046 & 0.143 & 0.150 & 0.033 & 0.048 & 0.196 & 0.203 \\
			0.10 & 0.051 & 0.074 & 0.143 & 0.154 & 0.058 & 0.081 & 0.202 & 0.214 \\
			0.20 & 0.075 & 0.108 & 0.134 & 0.149 & 0.087 & 0.117 & 0.180 & 0.201 \\
			0.30 & 0.084 & 0.124 & 0.123 & 0.142 & 0.101 & 0.130 & 0.150 & 0.177 \\
			0.40 & 0.088 & 0.129 & 0.114 & 0.135 & 0.105 & 0.129 & 0.127 & 0.158 \\
			0.50 & 0.087 & 0.125 & 0.110 & 0.130 & 0.102 & 0.122 & 0.118 & 0.150 \\
			0.60 & 0.087 & 0.116 & 0.108 & 0.127 & 0.094 & 0.112 & 0.127 & 0.154 \\
			0.70 & 0.074 & 0.100 & 0.110 & 0.127 & 0.082 & 0.097 & 0.151 & 0.168 \\
			0.80 & 0.061 & 0.082 & 0.111 & 0.124 & 0.068 & 0.080 & 0.177 & 0.186 \\
			0.90 & 0.042 & 0.057 & 0.104 & 0.113 & 0.046 & 0.056 & 0.193 & 0.198 \\
			0.95 & 0.027 & 0.037 & 0.092 & 0.098 & 0.029 & 0.037 & 0.183 & 0.186 \\
			0.99 & 0.009 & 0.013 & 0.066 & 0.068 & 0.011 & 0.014 & 0.128 & 0.130 \\
			\bottomrule
		\end{tabular}
		\caption{Per-level calibration errors on the extended grid, averaged over horizons and the 42 configurations with at least 50 forecast windows.}
		\label{tab:tails}
	\end{table}
	
	\paragraph{Number of Monte Carlo samples.}
	Table~\ref{tab:samples} recomputes all metrics from random subsets of \(S\in\{20,50,100,200,500\}\) samples for the three sample-based models. All metrics stabilize at \(S\approx100\), and moving to \(S=500\) changes them by at most 0.002. The large errors of Chronos-T5 and Sundial are therefore not Monte Carlo artifacts.
	
	\begin{table}[htb]
		\centering
		\small
		\begin{tabular}{@{} l r rrrr @{}} \toprule
			\textbf{Model} & \(S\) & \(\mathrm{PCE}_{\mathrm{pool}}\) & \(\mathrm{PCE}_{\mathrm{avg}}\) & \(\mathrm{QCE}_{\mathrm{avg}}\) & \(\Delta_C\) \\
			\midrule
			\cellcolor{cat1color!12}Moirai-Base & 20 & 0.0499 & 0.0734 & 0.1084 & 0.0350 \\
			\cellcolor{cat1color!12}Moirai-Base & 50 & 0.0463 & 0.0717 & 0.1038 & 0.0322 \\
			\cellcolor{cat1color!12}Moirai-Base & 100 & 0.0461 & 0.0717 & 0.1025 & 0.0308 \\
			\cellcolor{cat1color!12}Moirai-Base & 200 & 0.0462 & 0.0720 & 0.1021 & 0.0301 \\
			\cellcolor{cat1color!12}Moirai-Base & 500 & 0.0463 & 0.0722 & 0.1018 & 0.0296 \\
			\midrule
			\cellcolor{cat2color!12}Chronos-T5-Small & 20 & 0.1084 & 0.1249 & 0.1409 & 0.0160 \\
			\cellcolor{cat2color!12}Chronos-T5-Small & 50 & 0.1025 & 0.1201 & 0.1361 & 0.0160 \\
			\cellcolor{cat2color!12}Chronos-T5-Small & 100 & 0.1005 & 0.1186 & 0.1345 & 0.0158 \\
			\cellcolor{cat2color!12}Chronos-T5-Small & 200 & 0.0995 & 0.1180 & 0.1338 & 0.0158 \\
			\cellcolor{cat2color!12}Chronos-T5-Small & 500 & 0.0989 & 0.1175 & 0.1333 & 0.0158 \\
			\midrule
			\cellcolor{cat4color!12}Sundial-Base & 20 & 0.1489 & 0.1650 & 0.1849 & 0.0200 \\
			\cellcolor{cat4color!12}Sundial-Base & 50 & 0.1446 & 0.1611 & 0.1811 & 0.0200 \\
			\cellcolor{cat4color!12}Sundial-Base & 100 & 0.1430 & 0.1596 & 0.1796 & 0.0200 \\
			\cellcolor{cat4color!12}Sundial-Base & 200 & 0.1421 & 0.1589 & 0.1789 & 0.0200 \\
			\cellcolor{cat4color!12}Sundial-Base & 500 & 0.1416 & 0.1584 & 0.1784 & 0.0200 \\
			\bottomrule
		\end{tabular}
		\caption{Sensitivity of calibration metrics to the number of samples \(S\) (42 configurations).}
		\label{tab:samples}
	\end{table}
	
	\paragraph{Scale differences between series.}
	GIFT-Eval datasets combine series of very different magnitudes, so conditioning on raw predicted values could partly condition on the identity of the series. We therefore recompute QCE after standardizing every predicted quantile by the mean and standard deviation of its context window, which mirrors the instance normalization of TSFMs; coverage indicators are unchanged. Averaged over the 588 model--dataset pairs with at least 50 windows, \(\mathrm{QCE}_{\mathrm{avg}}\) changes from 0.112 to 0.110 and \(\Delta_C\) from 0.026 to 0.023, so value dependence is not an artifact of scale differences.
	
	\paragraph{Horizon length.}
	Table~\ref{tab:horizon} compares the short-, medium-, and long-term tasks of the 21 configurations that support all three. The worst-horizon error more than doubles from short to long horizons, and the horizon gap nearly triples, so longer horizons contain increasingly heterogeneous step-wise calibration that pooled coverage hides.
	
	\begin{table}[htb]
		\centering
		\small
		\begin{tabular}{@{} l r rrrr @{}} \toprule
			\textbf{Task} & \textbf{avg.\ \(H\)} & \(\mathrm{PCE}_{\mathrm{pool}}\) & \(\mathrm{PCE}_{\mathrm{avg}}\) & \(\mathrm{PCE}_{\max}\) & \(\Delta_H\) \\
			\midrule
			Short & 49 & 0.056 & 0.073 & 0.130 & 0.017 \\
			Medium & 491 & 0.085 & 0.125 & 0.260 & 0.040 \\
			Long & 737 & 0.097 & 0.144 & 0.297 & 0.047 \\
			\bottomrule
		\end{tabular}
		\caption{Calibration errors by forecasting task on the 21 configurations with short-, medium-, and long-term tasks (mean over models and configurations).}
		\label{tab:horizon}
	\end{table}
	
	\paragraph{Outcomes with point masses.}
	Four configurations have more than 40\% of their targets at a single value (\texttt{car\_parts}, \texttt{solar/10T}, \texttt{solar/H}, \texttt{temperature\_rain}). As explained in Appendix~\ref{sec:calib_computation}, even a perfect forecast can show coverage errors there. These configurations produce the largest raw conditioning gaps, but excluding all configurations with more than 5\% of their targets at a single value changes the mean \(\Delta_C\) only from 0.035 to 0.033.

	\clearpage
	\input{checklist.tex}
	
\end{document}

%% file: checklist.tex
\section*{NeurIPS Paper Checklist}



\begin{enumerate}

\item {\bf Claims}
    \item[] Question: Do the main claims made in the abstract and introduction accurately reflect the paper's contributions and scope?
    \item[] Answer: \answerYes{},
    \item[] Justification: All contributions relate to sections in the theoretical and empirical analysis.

\item {\bf Limitations}
    \item[] Question: Does the paper discuss the limitations of the work performed by the authors?
    \item[] Answer: \answerYes{},
    \item[] Justification: We discuss the limitation in the final section.

\item {\bf Theory assumptions and proofs}
    \item[] Question: For each theoretical result, does the paper provide the full set of assumptions and a complete (and correct) proof?
    \item[] Answer: \answerYes{}, 
    \item[] Justification: We provide complete proofs in the appendix.

    \item {\bf Experimental result reproducibility}
    \item[] Question: Does the paper fully disclose all the information needed to reproduce the main experimental results of the paper to the extent that it affects the main claims and/or conclusions of the paper (regardless of whether the code and data are provided or not)?
    \item[] Answer: \answerYes{}, 
    \item[] Justification: We documented detailed experimental details in the appendix and provide supplementary code. 

\item {\bf Open access to data and code}
    \item[] Question: Does the paper provide open access to the data and code, with sufficient instructions to faithfully reproduce the main experimental results, as described in supplemental material?
    \item[] Answer: \answerYes{},
    \item[] Justification: We provide the underlying code via a public GitHub repository.

\item {\bf Experimental setting/details}
    \item[] Question: Does the paper specify all the training and test details (e.g., data splits, hyperparameters, how they were chosen, type of optimizer) necessary to understand the results?
    \item[] Answer: \answerYes{}, 
    \item[] Justification: Details can be derived from the appendix and the code.

\item {\bf Experiment statistical significance}
    \item[] Question: Does the paper report error bars suitably and correctly defined or other appropriate information about the statistical significance of the experiments?
    \item[] Answer: \answerYes{},
    \item[] Justification: We provide tables with standard error when applicable.

\item {\bf Experiments compute resources}
    \item[] Question: For each experiment, does the paper provide sufficient information on the computer resources (type of compute workers, memory, time of execution) needed to reproduce the experiments?
    \item[] Answer: \answerYes{} 
    \item[] Justification: The compute ressources used are documented in the appendix.
    
\item {\bf Code of ethics}
    \item[] Question: Does the research conducted in the paper conform, in every respect, with the NeurIPS Code of Ethics \url{https://neurips.cc/public/EthicsGuidelines}?
    \item[] Answer: \answerYes{} 
    \item[] Justification: NA.

\item {\bf Broader impacts}
    \item[] Question: Does the paper discuss both potential positive societal impacts and negative societal impacts of the work performed?
    \item[] Answer: \answerNA{}
    \item[] Justification: NA
    
\item {\bf Safeguards}
    \item[] Question: Does the paper describe safeguards that have been put in place for responsible release of data or models that have a high risk for misuse (e.g., pre-trained language models, image generators, or scraped datasets)?
    \item[] Answer: \answerNA{}
    \item[] Justification: NA

\item {\bf Licenses for existing assets}
    \item[] Question: Are the creators or original owners of assets (e.g., code, data, models), used in the paper, properly credited and are the license and terms of use explicitly mentioned and properly respected?
    \item[] Answer: \answerYes{}
    \item[] Justification: NA

\item {\bf New assets}
    \item[] Question: Are new assets introduced in the paper well documented and is the documentation provided alongside the assets?
    \item[] Answer: \answerNA{} 
    \item[] Justification: NA

\item {\bf Crowdsourcing and research with human subjects}
    \item[] Question: For crowdsourcing experiments and research with human subjects, does the paper include the full text of instructions given to participants and screenshots, if applicable, as well as details about compensation (if any)? 
    \item[] Answer: \answerNA{} 
    \item[] Justification: NA

\item {\bf Institutional review board (IRB) approvals or equivalent for research with human subjects}
    \item[] Question: Does the paper describe potential risks incurred by study participants, whether such risks were disclosed to the subjects, and whether Institutional Review Board (IRB) approvals (or an equivalent approval/review based on the requirements of your country or institution) were obtained?
    \item[] Answer: \answerNA{} 
    \item[] Justification: NA

\item {\bf Declaration of LLM usage}
    \item[] Question: Does the paper describe the usage of LLMs if it is an important, original, or non-standard component of the core methods in this research? Note that if the LLM is used only for writing, editing, or formatting purposes and does \emph{not} impact the core methodology, scientific rigor, or originality of the research, declaration is not required.
    \item[] Answer: \answerNA{}
    \item[] Justification: NA

\end{enumerate}